\documentclass[11pt]{article}

\usepackage{acl}

\usepackage{times}
\usepackage{latexsym}
\usepackage[T1]{fontenc}
\usepackage[utf8]{inputenc}
\usepackage{microtype}
\usepackage{inconsolata}
\usepackage{graphicx}
\usepackage{booktabs}
\usepackage{multirow}
\usepackage{array}
\usepackage[table]{xcolor}
\usepackage{colortbl}
\usepackage{amsmath}
\usepackage{float}
\usepackage{placeins}
\usepackage{hyperref}
\usepackage{tabularx}
\usepackage{listings}
\usepackage{comment}
\usepackage{enumitem}

\lstdefinestyle{prompt}{
  basicstyle=\small\ttfamily,
  breaklines=true,
  breakatwhitespace=true,
  breakindent=0pt,
  breakautoindent=false,
  backgroundcolor=\color{gray!7},
  frame=none,
  xleftmargin=0pt,
  xrightmargin=0pt,
  aboveskip=4pt,
  belowskip=8pt,
  keepspaces=true,
}

\definecolor{rowgray}{HTML}{EFEFEF}
\definecolor{cellred}{HTML}{FFC9C9}

\graphicspath{{images/}}

\title{Rating the Pitch, Not the Product: User Evaluations of LLMs Reflect Expectations More Than Performance}

\author{
  \textbf{Robert Morabito\textsuperscript{1}},
  \textbf{Tyler McDonald\textsuperscript{1}},
  \textbf{Charitra Viswanath\textsuperscript{2}},
  \\
  \textbf{Angel Hsing-Chi Hwang\textsuperscript{3}},
  \textbf{Susanne Gaube\textsuperscript{4}},
  \textbf{Jad Kabbara\textsuperscript{5}},
  \textbf{Ali Emami\textsuperscript{2}}
  \\
  \textsuperscript{1}Brock University, Saint Catharines, Canada,
  \textsuperscript{2}Emory University, Atlanta, USA
  \\
  \textsuperscript{3}University of Southern California, Los Angeles, USA
  \textsuperscript{4}University College London, London, UK
  \\
  \textsuperscript{5}Massachusetts Institute of Technology, Cambridge, USA
\\
}

\begin{document}
\maketitle

\begin{abstract}
Imagine two users interact with the same LLM. One is told it is the cutting-edge flagship; the other, an older, weaker model. They walk away with markedly different ratings of its usefulness and intelligence, yet they used the \textit{same model}. In a controlled study, 162 participants each used one of six LLMs from two families across three collaborative tasks, after first viewing a landing page that matched, overstated, or understated their model's true capability. This pre-interaction \textit{framing} persisted through use. Oversold users rated the model more favorably than Undersold users before interaction; afterward, both groups had updated toward reality, but the impression gap did not fully close. Framing also changed how users worked with the model (with Oversold users prompting more directively, Undersold users more collaboratively), but the quality of what they produced together depended only on the model's true capability, not on what users were told. Participants' change in model impressions after use, measured across two impression measures, was \textit{not} predicted by task performance ($\beta = -0.01$ and $0.11$, both n.s.), but by whether the model met users' expectations ($\beta = 0.47$ and $0.50$, both $p < .001$) and how confident they felt working with it ($\beta = 0.47$ and $0.36$, both $p < .001$). After interaction, users are still rating the \textit{pitch}, not the \textit{product}: user-elicited LLM evaluations, including the preference data driving public leaderboards, measure expectation management at least as much as the model itself.
\end{abstract}

% Imagine two users interact with the same LLM. One has been told it is the cutting-edge flagship model; the other, an older, weaker model. They walk away with markedly different ratings of its usefulness and intelligence, yet they used the \textit{same model}. In a controlled study, 162 participants each used one of six LLMs from two families across three collaborative tasks, after first viewing a landing page that matched, overstated, or understated their model's true capability. This pre-interaction \textit{framing} shifted user opinions and interaction behavior while task performance did not. Oversold users rated the model more favorably and used more directive prompting, while Undersold users wrote longer, more collaborative prompts. The quality of what users and the model produced together depended only on the model's true capability, not on what users were told. Participants' change in model impressions after use, measured across two impression measures, was \textit{not} predicted by task performance ($\beta = -0.01$ and $0.11$, both n.s.), but by whether the model met users' expectations ($\beta = 0.47$ and $0.50$, both $p < .001$) and how confident they felt working with it ($\beta = 0.47$ and $0.36$, both $p < .001$). After interaction, users are still rating the \textit{pitch}, not the \textit{product}: user-elicited LLM evaluations, including the preference data driving public leaderboards, measure expectation management at least as much as the model itself.

\section{Introduction}
\label{sec:intro}

\begin{figure*}[t]
  \centering
  \includegraphics[width=\linewidth]{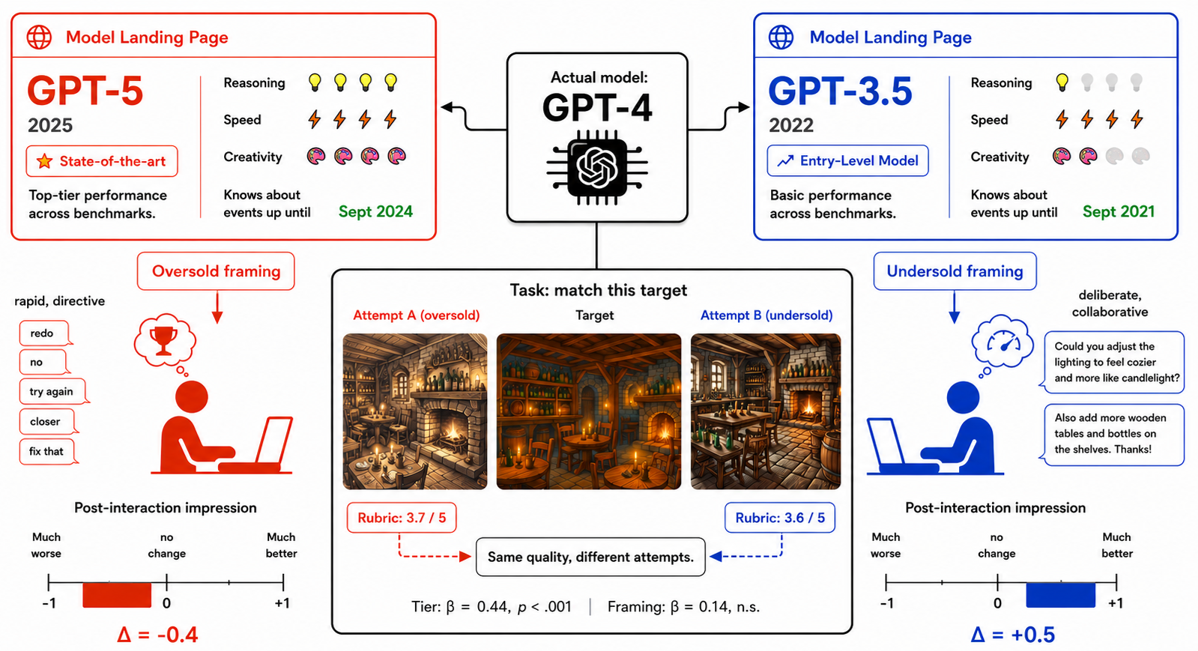}
  \caption{\textbf{Same model, different framings, different user experience.} Illustrated with GPT-4 framed as either state-of-the-art (Oversold) or entry-level (Undersold) using a real task and outputs from the study; the full study spans six models across GPT and Claude families. Framing changed user interaction behavior and post-interaction impressions, but not task performance.}
  \label{fig:overview}
\end{figure*}

Most users meet an LLM before they \textit{actually} meet it. They might have seen a benchmark score, a launch post, a leaderboard ranking \citep{10.5555/3692070.3692401}, or heard a colleague's recommendation. These pre-interaction signals may have lasting consequences for how users evaluate, adopt, and trust the systems they go on to use. Do initial expectations fade as users accumulate experience, or do they persist? Do they only affect impressions, or also how users interact with the model and what they produce together? And when impressions do change, what drives that change: the quality of the output, or something else about the experience?

Pre-interaction expectations are well-known drivers of technology evaluation, adoption, and abandonment \citep{10.2307/3250921, Budhathoki03052024}, and post-interaction impressions shape satisfaction often independently of objective performance \citep{blut2021understanding, jtaer21040122}. As LLMs become embedded in workflows where mismatched expectations can drive misuse or misplaced trust \citep{doi:10.1126/science.adh2586}, how expectations evolve into impressions through use is consequential.

Work in human-AI interaction has begun to ask these questions. The \textit{framing} of an AI system (the description users encounter before use) shifts judgments of trustworthiness \citep{pataranutaporn2023influencing}, misjudged capability changes what users produce \citep{10.1145/3706598.3713886}, and passive rather than collaborative use changes their sense of ownership over the output \citep{lee2026relying}. No prior work, however, has followed users across iterative LLM use to ask whether these effects persist or generalize to behavior and output, and what sustains them. For LLMs, where users engage in multi-turn collaboration and capability is signaled through public benchmarks \citep{10.1145/3290605.3300641}, marketing, and word of mouth, these questions matter for individual users and the preference-based evaluation pipelines that aggregate their judgments.

% None, however, have followed users across iterative LLM use to ask whether these effects persist, generalize to behavior and output, or what sustains them.

We ran a controlled between-subjects study with 162 participants, each randomly assigned a \textbf{source} model (the model actually serving the participant) and a \textbf{shown} model (the model described to them), which determines framing condition (\textit{Oversold}, \textit{Matched}, and \textit{Undersold}), from two model families (GPT \& Claude). Each first saw a landing page introducing a model from their assigned family, which did not always match the model actually serving them in the backend (Appendix Figure~\ref{fig:appendix-framing}): Matched saw their true model, Oversold saw a higher-tier model (e.g., they were \textit{using} GPT-3.5 but were \textit{shown} GPT-5), and Undersold saw a lower-tier model (e.g., they were \textit{using} GPT-5 but were \textit{shown} GPT-3.5). All completed three collaborative tasks (image generation, outreach message writing, and acronym building). We measured impressions before and after the session, task-specific confidence around each task, interaction behavior from conversation logs, and output quality. Figure~\ref{fig:overview} summarizes the design and central results.

%The manipulation worked as intended: before interaction, Oversold participants rated the model's expected usefulness and intelligence more highly than Undersold ones. After use, both groups updated toward reality (Undersold upward, Oversold downward), but the impression gap did not fully close. Framing also reshaped how users interacted, though not what they produced; what predicted impression change was whether the model met their expectations and how their confidence shifted, not the output itself.

\noindent We establish three findings:
\begin{itemize}[itemsep=1pt, topsep=1pt, parsep=1pt, leftmargin=*]

    \item \textbf{The impression gap set by framing does not close with use.} Oversold participants entered with higher impressions and Undersold with lower; after three tasks of hands-on work, both groups updated towards reality, but the gap did not fully close, though expectation-disconfirmation accounts predict that experience should wash the initial framing out (\S\ref{sec:rq1}). 
    
    %\item \textbf{Pre-interaction framing of an LLM persistently shapes user impressions.} Users in the \textit{Oversold} condition rated models more favorably before use and less favorably after; the reverse held for those in the \textit{Undersold} condition (\S\ref{sec:rq1}).

    \item \textbf{Framing does not change the quality of what users and the model produce together, only how they work to produce it.} Under different framings, the same model produced different interaction behavior: Oversold users prompted more directively, Undersold users more collaboratively, with the divergence largest on the acronym task, however task performance did not vary with framing (\S\ref{sec:rq2}, \S\ref{sec:rq3}).
    
    \item \textbf{Impression change is driven by experience, not output.} What predicts how users' impressions of an LLM shift after using it is not their task performance ($\beta{=}-0.01$ and $0.11$, both n.s.), but whether the model met their expectations and how their task-specific confidence (i.e., their \textit{self-efficacy}) changed ($\beta{=}0.36$ to $0.50$, all $p<.001$) (\S\ref{sec:rq4}).
\end{itemize}

%We situate these findings against prior work on framing effects, subjective versus objective AI evaluation, and human interaction behavior in \S\ref{sec:related}.

The implications are broad. Users' impressions of a model are partly set before any prompt is sent. For organizations deploying AI tools, \textit{how} a tool is introduced may matter as much as the tool itself. For developers and benchmark designers, expectation-setting signals (benchmarks, marketing, interfaces) appear to drive user impressions more than capability differences users detect during use, particularly for open-ended tasks (e.g., those involving creativity, expression, and opinion). For evaluation research, the concern is one of measurement validity: preference data elicited from users may measure expectation management at least as much as model capability.

\section{Methodology}
\label{sec:method}

To test whether pre-interaction framing of an LLM's capability persists into use, and what drives impressions if it does, we crossed three source tiers with three shown tiers, across two model families: GPT \cite{ouyang2022instructgpt, openai2023gpt4, openai2025gpt5} and Claude \cite{anthropic2024claude3, anthropic2024claude35, anthropic2025claude4}. Framing (\textit{Oversold}, \textit{Matched}, \textit{Undersold}) is the relation between the two. 162 participants completed three collaborative tasks of varying objectivity; we measured impressions of the model before and after the session, self-efficacy before and after each task, expectations met after each task, and the quality of the final output, while tracking behavioral metrics.

% To test whether pre-interaction framing of an LLM's capability persists into use, and what drives impressions if it does, we crossed three framing levels (\textit{Oversold}, \textit{Matched}, \textit{Undersold}) with three capability tiers (Bottom, Middle, Top) across two model families: GPT \cite{ouyang2022instructgpt, openai2023gpt4, openai2025gpt5} and Claude \cite{anthropic2024claude3, anthropic2024claude35, anthropic2025claude4}. 162 participants completed three collaborative tasks of varying objectivity; we measured impressions before and after, behavior during, and the quality of the final output.

\subsection{Capability Framing}
\label{sec:design}

\begin{table}[t]
\centering
\small
\resizebox{\linewidth}{!}{
\begin{tabular}{llll}
    \toprule
    \textbf{\begin{tabular}[c]{@{}l@{}}Source tiers (rows) vs. \\ Shown tiers (columns)\end{tabular}} & \textbf{Bottom} & \textbf{Middle} & \textbf{Top} \\
    \midrule
    \textbf{Bottom} & \cellcolor[HTML]{C0F8BF}Matched   & \cellcolor[HTML]{FFCCC9}Oversold  & \cellcolor[HTML]{FFCCC9}Oversold \\
    \textbf{Middle} & \cellcolor[HTML]{D4E6FF}Undersold & \cellcolor[HTML]{C0F8BF}Matched   & \cellcolor[HTML]{FFCCC9}Oversold \\
    \textbf{Top}    & \cellcolor[HTML]{D4E6FF}Undersold & \cellcolor[HTML]{D4E6FF}Undersold & \cellcolor[HTML]{C0F8BF}Matched  \\
    \bottomrule
\end{tabular}
}
\vspace{-1mm}
\caption{Framing condition, determined by the relation between the model serving the participant (source tier) and the model described on the landing page (shown tier). Each combination was run in both model families.}
\label{tab:design}
\end{table}

Each participant was introduced to their assigned model through a landing page before any interaction began (Appendix Figure~\ref{fig:appendix-framing}). The page displayed the model's name, release year, knowledge cutoff, and three capability scores (reasoning, speed, and creativity) on a 1--4 scale, alongside a comparison against other models in its family. The reasoning and creativity scores were derived from the model's percentile rank on LMArena's Hard Prompts and Creative Writing leaderboards, respectively,\footnote{https://arena.ai/leaderboard/} at the time of the study; the speed score was derived from the latency metric from each provider's model overview page.\footnote{Anthropic: \href{https://platform.claude.com/docs/en/about-claude/models/overview}{Claude models overview}; OpenAI: \href{https://developers.openai.com/api/docs/models}{OpenAI models overview}} The capability tier shown did not always correspond to the model actually served by the backend.

Each family had three models at capability tiers ordered by public benchmark rankings at the time of data collection \citep{10.5555/3692070.3692401}. Table~\ref{tab:models} lists the six source models and the shown names within each family, where every source model was paired with every shown tier, yielding nine source-by-shown combinations per family and 18 across the full design. The framing condition follows from this pairing rather than being assigned separately: a participant is Oversold when the shown tier sits above the source tier, Undersold when it sits below, and Matched when the two agree (Table~\ref{tab:design}). Nine participants were assigned to each combination, for $N{=}162$, balanced at 54 per framing condition, 81 per family, and 27 per source model. These numbers were determined by a pilot of 18 participants (one per source-by-shown cell) and a Monte Carlo power analysis targeting 80\% power at $\alpha = 0.05$.

Testing whether output quality tracks real capability while impressions track the description requires true and described capability to vary independently. Three tiers per family supply that variation, and repeating the manipulation in a second family checks that the results are not specific to one lineage.

The pairing also creates a distance between source and shown tier, so we use framing extent (source tier minus shown tier, $-2$ to $+2$) to test for dose-response effects.

%We also analyzed a framing extent metric (source tier minus shown tier, range −2 to +2) to test for dose-response effects.

\begin{table}[t]
  \centering
  \small
\resizebox{0.85\linewidth}{!}{
  \begin{tabular}{lll}
    \toprule
    \textbf{Tier} & \textbf{Source model} & \textbf{Shown name} \\
    \midrule
    \multicolumn{3}{l}{\emph{GPT family}} \\
    Bottom & \texttt{gpt-3.5-turbo-0125}     & GPT-3.5 \\
    Middle & \texttt{gpt-4-0125-preview}     & GPT-4 \\
    Top    & \texttt{gpt-5-2025-08-07}       & GPT-5 \\
    \addlinespace[2pt]
    \multicolumn{3}{l}{\emph{Claude family}} \\
    Bottom & \texttt{claude-3-haiku-20240307}    & Claude 3 \\
    Middle & \texttt{claude-3-5-haiku-20241022}  & Claude 3.5 \\
    Top    & \texttt{claude-sonnet-4-20250514}   & Claude 4 \\
    \bottomrule
  \end{tabular}}
    \vspace{-1mm}
  \caption{Source models used in the study. Each source model was paired with each shown name in its family across participants, defining the framing manipulation.}
  \label{tab:models}
\end{table}

\subsection{Tasks}
\label{sec:tasks}

Participants completed three tasks in the same order: image generation, outreach message writing, and acronym building. 

\paragraph{Image generation.} Participants worked with the model to generate an image resembling the target image as closely as possible (Appendix Figure~\ref{fig:appendix-target}). DALL-E 3 rendered every image, with the source model refining the participant's request into the prompt, so differences between source tiers enter this task only through prompt quality.

\paragraph{Outreach message writing.} The task was to draft a persuasive outreach message for a sales role application (Appendix Table~\ref{tab:appendix-jobdesc}). The message had to fulfill several requirements: (a) reference the job description, (b) be signed ``J. Doe'', (c) be addressed to ``Dr. Rogers'', and (d) read as if written by a human rather than an AI tool.
\paragraph{Acronym building.} The goal was to generate humorous acronyms from a set of letters (e.g., `\textbf{PUGT}' → `\textbf{P}rolific \textbf{U}sers \textbf{G}et \textbf{T}ired'). They worked with three letter sets: (a) B L M P F, (b) C R T W, and (c) Y O L D. With no objectively correct answer or specific requirements to fulfill, quality depended entirely on the participant's creative direction in collaboration with the model.

All tasks required sustained, iterative collaboration with the model rather than single-turn prompting. The tasks were chosen to reflect varying levels of objectivity: the first has a fixed external target, the second is anchored to a written brief, and the third is fully open-ended and subjective. This spread allows us to observe whether framing effects vary with how much the user's own creative input shapes what counts as a good output.

Participants were told each task required at least five minutes of interaction and the highest-scoring submission per task would earn a \$10 bonus.

\subsection{Measures}
\label{sec:measures}

\paragraph{Impressions.} We used the Unified Theory of Acceptance and Use of Technology (UTAUT) Performance Expectancy and Effort Expectancy subscales \citep{10.2307/30036540}, which capture usefulness and ease of use respectively (Appendix Table~\ref{tab:appendix-utaut}), and the Godspeed Perceived Intelligence subscale \citep{bartneck2009measurement}, which captures perceived intelligence (Appendix Table~\ref{tab:appendix-godspeed}). Both were administered before the study (future tense) and after (past tense).

\paragraph{Self-efficacy.} Self-efficacy refers to a person's belief in their ability to perform a task \citep{doi:10.1177/109442810141004}. We adapted the New General Self-Efficacy scale into eight task-specific items, completed before and after each task, to capture how participants' self-efficacy in working with the model changed across the study (Appendix Table~\ref{tab:appendix-ngse}).

\paragraph{Expectations met.} After each task, participants rated if the model fell short, met, or exceeded their expectations for that task on a three-point scale.

\paragraph{Behavior.} From each participant's task conversation logs, we extracted 17 behavioral metrics, covering message counts and lengths, keystroke and backspace counts, edit events, session and idle time, and time between messages (full list in Appendix Table~\ref{tab:appendix-behavior}). Each user message was classified as \emph{collaborative} or \emph{directive} by \texttt{claude-haiku-4-5}, using a prompt grounded in \citet{searle_1976}'s illocutionary taxonomy and \citet{10.1007/978-3-642-85098-1_5}'s work on collaborative engagement (prompt and method in Appendix~\ref{sec:appendix-colab}). Human annotators validated the classifier on a stratified sample of messages, and its agreement with them was fair against their own moderate agreement (Appendix~\ref{sec:appendix-classifier-validation}).

\paragraph{Output quality.} 
Participants selected and submitted one final product per task, which was scored by GPT-5 against a task-specific rubric on a 1–5 scale across four dimensions, averaged over three scoring runs (Rubric in Appendix Table~\ref{tab:appendix-rubric}; full prompt and method in Appendix~\ref{sec:appendix-judge}). Human raters validated the judge on a subset of submissions, where its correlation to humans across tasks was similar to humans' correlation to each other (Appendix~\ref{sec:appendix-validation}). 

Examples of high- and low-scoring submissions for image generation are shown in Figures~\ref{fig:appendix-img-best} and~\ref{fig:appendix-img-worst}, for outreach messages in Table~\ref{tab:appendix-msg-examples}, and for acronyms in Table~\ref{tab:appendix-acro-examples}, of Appendix~\ref{sec:appendix-rubrics-examples}.

\subsection{Participants and Procedure}
\label{sec:procedure}

\paragraph{Recruitment.} We recruited 162 US-based workers through \href{https://www.prolific.com/}{Prolific}, all of whom were fluent in English, had a 98--100\% approval rate, and had completed more than 100 prior studies. Alongside the study measures we collected each participant's chatbot use frequency and self-reported comfort with chatbots. Assignment to the 18 source-by-shown cells was managed by an SQL database that atomically reserved an open cell for each incoming participant, preventing race conditions and maintaining the balanced design across the full data collection period. All study procedures were approved by the authors' institutional research ethics board prior to data collection, and all participants provided informed consent before beginning the study.

\paragraph{Procedure.} The study was administered across Prolific, Tally, and a custom web-based chat interface powered by the source LLM (Appendix Figure~\ref{fig:appendix-interface}). After consent and pre-study UTAUT and Godspeed surveys, participants viewed the landing page (manipulation) and had to pass a 3-item manipulation check. They then completed the three tasks in sequence, each bracketed by the self-efficacy scale, and rated whether the model met their expectations after each. Post-study UTAUT and Godspeed surveys closed the session.

\paragraph{Data integrity.} To detect automated or low-effort participation, we implemented layered checks. Each task page contained a honeypot button invisible to human users but detectable by browser automation agents, hidden DOM elements with task-specific seeded keywords later searched in conversation logs, and browser-environment checks flagging known indicators of automation. Three attention check items were embedded across the surveys (Appendix~\ref{sec:appendix-surveys}). All submissions were then reviewed manually, blind to framing condition and before any outcome analysis; ten participants were removed and replaced through new recruitment, preserving the balanced $N{=}162$ (criteria and counts in Appendix~\ref{sec:appendix-filtering}).

\subsection{Analysis}
\label{sec:analysis}

\paragraph{Codings and outcomes.} Analyses were conducted in R. Framing was coded $\{-1, 0, +1\}$ (Under/Matched/Over) and model tier $\{-1, 0, +1\}$ (Bottom/Middle/Top); framing extent was used as an integer covariate ($-2$ to $+2$). Impression scores are mean item ratings for the UTAUT and Godspeed respectively; impression change scores are within-participant differences, each person's own post composite minus their own pre composite (i.e., $UTAUT_{post} - UTAUT_{pre}$ and $Godspeed_{post} - Godspeed_{pre}$). Self-efficacy change is computed per task ($SE_{post} - SE_{pre}$) and averaged across the three tasks. Overall performance is the equal-weighted mean of the three task-level judge scores.

\begin{table}[t]
  \centering
  \small
  \begin{tabular}{llcc}
    \toprule
    \textbf{Measure} & \textbf{Condition} & \textbf{Mean} & \textbf{SD} \\
    \midrule
    \multirow{3}{*}{Pre-UTAUT}    & Oversold  & 4.21 & 0.65 \\
                                  & Matched   & 4.05 & 0.73 \\
                                  & Undersold & 3.96 & 0.54 \\
    \addlinespace[2pt]
    \multirow{3}{*}{Pre-Godspeed} & Oversold  & 5.46 & 1.05 \\
                                  & Matched   & 5.33 & 1.04 \\
                                  & Undersold & 4.79 & 0.97 \\
    \bottomrule
  \end{tabular}
  \vspace{-1mm}
\caption{Pre-interaction UTAUT (1--5; higher = more useful and easier to use) and Godspeed Perceived Intelligence (1--7; higher = more intelligent and competent) by framing condition. $N{=}54$ per condition.}
  \label{tab:rq1-descriptives}
\end{table}

\begin{figure}[t]

  \centering
  \includegraphics[width=0.75\linewidth]{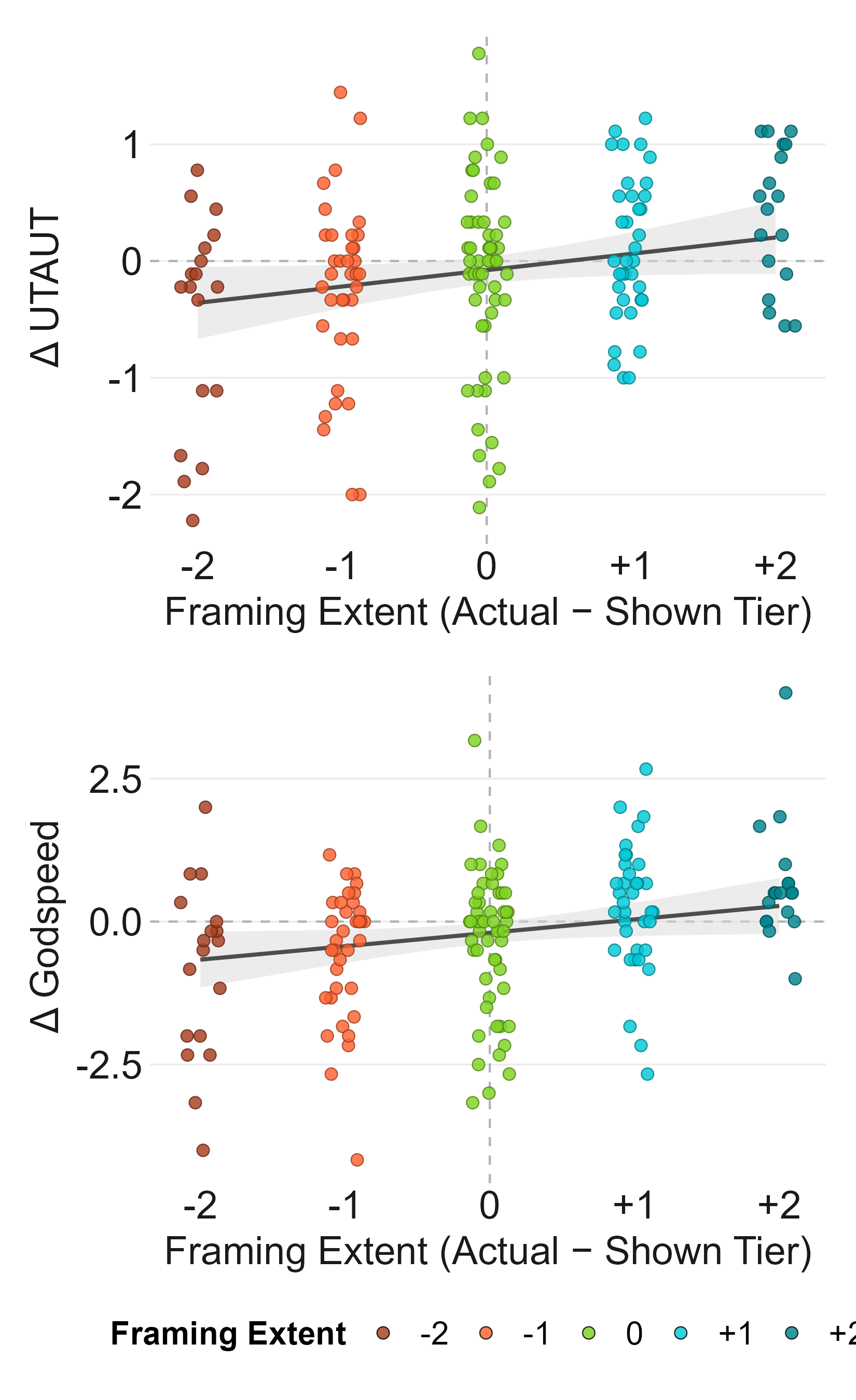}
  \vspace{-2mm}
  \caption{OLS regression of impression change on framing extent (source tier $-$ shown tier), with source tier held at its mean. Points are jittered participants; band is the 95\% CI on the regression line.}
  \label{fig:framing-extent}
\end{figure}

\paragraph{Statistical tests.} We used independent-samples $t$-tests with Cohen's $d$ for pre-study impression contrasts (\S\ref{sec:rq1}). Bar-chart figures show group means with $95\%$ CIs. Ordinary Least Squares (OLS) regression was used for the framing-extent dose-response model (\S\ref{sec:rq1}, Figure~\ref{fig:framing-extent}, jointly with source tier) and the performance regression on framing and model tier (\S\ref{sec:rq3}). For \S\ref{sec:rq4}, we ran six bivariate OLS regressions of each impression-change outcome on three predictors: task performance, self-efficacy change, and expectations met.

\paragraph{Multiple comparisons.} All p-values are reported uncorrected. In place of an $\alpha$-adjustment, each effect of theoretical interest is tested on two impression measures drawn from different validated instruments, and in \S\ref{sec:rq4} against two independent experiential predictors; all central effects replicate across both (more information in \S\ref{sec:limit}).

%All $p$-values are reported uncorrected. Our safeguard against false positives is built into the design: each effect of theoretical interest is tested independently on two impression measures (UTAUT and Godspeed) that draw from different validated instruments, and in \S\ref{sec:rq4} against two independent experiential predictors (self-efficacy change and expectations met). All central effects in \S\ref{sec:rq1} and \S\ref{sec:rq4} replicate across both impression outcomes and, where applicable, both predictors, providing design-level robustness rather than relying on $\alpha$-adjustment.

\begin{figure}[t]
  \centering
  \includegraphics[width=0.95\linewidth]{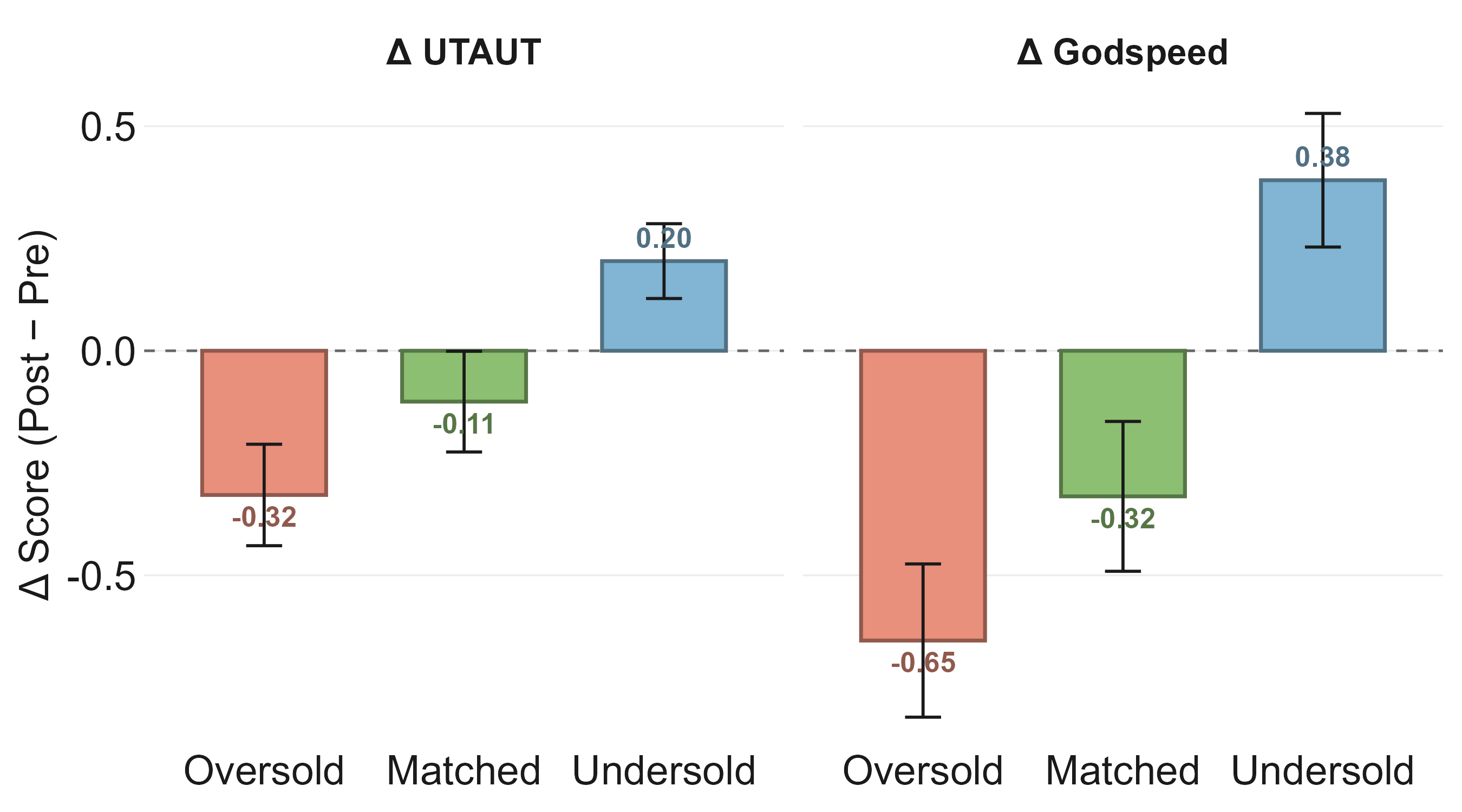}
  \vspace{-1mm}
  \caption{Mean impression change ($\Delta = \text{post} - \text{pre}$) on UTAUT (left) and Godspeed (right) by framing condition. Error bars are $\pm 1$ bootstrap SE ($B{=}5000$).}
  \label{fig:impression-change}
  \vspace{-1mm}
\end{figure}

\section{Results}
\label{sec:results}

\begin{figure*}[t]
  \centering
  \includegraphics[width=0.75\linewidth]{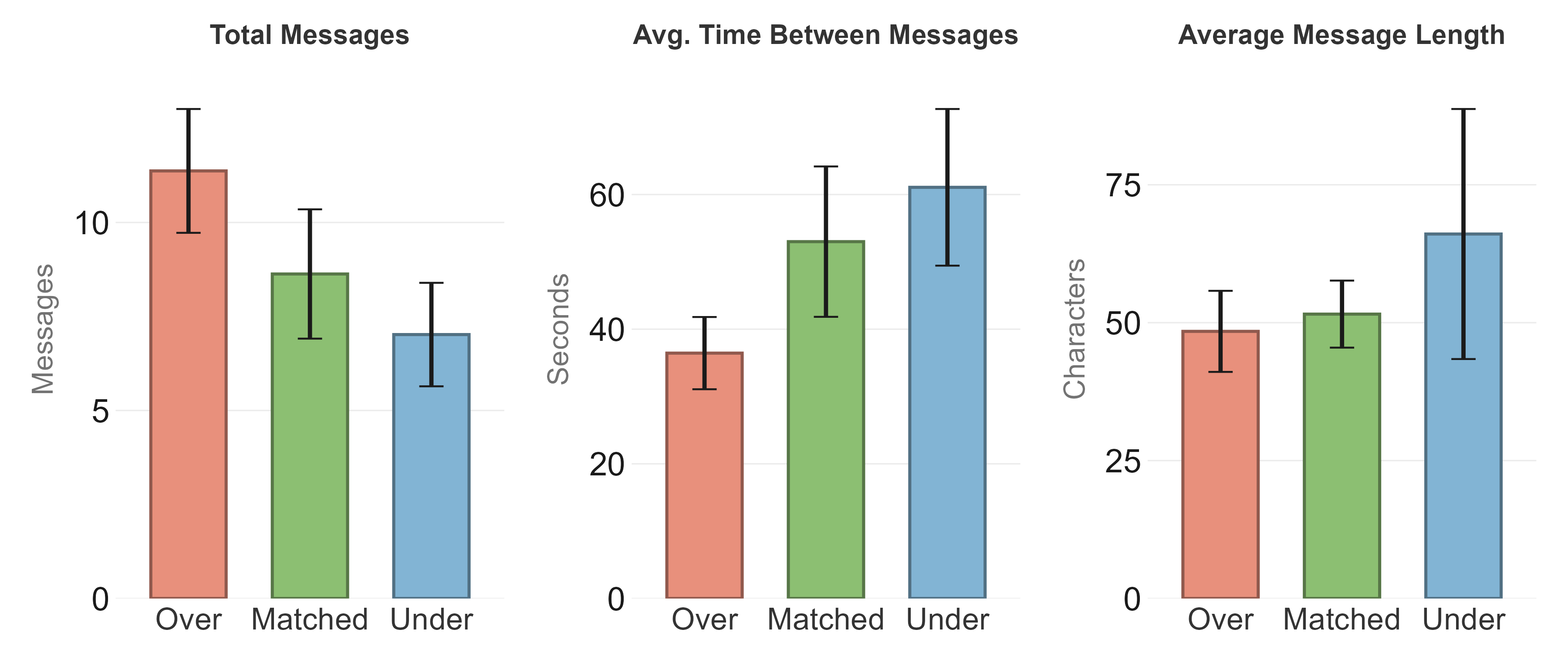}
  \vspace{-2mm}
  \caption{Three engagement metrics on the acronym task by framing condition (mean $\pm 95\%$ CI). From left to right: total messages sent, mean time between messages, and mean message length.}
  \label{fig:rq2-engagement}
\end{figure*}

\subsection{Framing sets an impression gap that use does not close}
\label{sec:rq1}

Before interaction, framing produced a clear ranking of impressions. Oversold participants entered with the highest ratings (UTAUT $M{=}4.21$, Godspeed $M{=}5.46$) and Undersold the lowest ($M{=}3.96$ and $M{=}4.79$), with Matched in between (Table~\ref{tab:rq1-descriptives}). The contrast between Oversold and Undersold was significant on both pre-UTAUT ($p{=}0.029$, $d{=}0.43$) and pre-Godspeed ($p{=}0.0009$, $d{=}0.66$) (pairwise contrasts in Appendix Table~\ref{tab:appendix-ttests}). The effect was larger on Godspeed than on UTAUT, indicating that perceived intelligence is more sensitive to framing than expected performance or effort.

Expectation-disconfirmation accounts predict that users revise their expectations toward reality as they accumulate experience \citep{oliver1980cognitive, 10.2307/3250921}, so three tasks of hands-on collaboration should have closed the gap that framing opened. It narrowed but did not close: ratings shifted toward the models' actual performance yet remained differentiated across conditions (Figure~\ref{fig:impression-change}). Undersold participants updated upward, Oversold updated downward, and Matched drifted slightly downward, suggesting that even an accurately described model can mildly underwhelm. The pattern held for both scales (Appendix Figure~\ref{fig:appendix-prepost}).

Impression change followed a dose-response relationship with framing extent: controlling for source tier, both $\Delta_{\mathrm{UTAUT}}$ and $\Delta_{\mathrm{Godspeed}}$ rose monotonically (Figure~\ref{fig:framing-extent}). The more Undersold the framing, the larger the upward update; the more Oversold, the larger the downward update.

The size of these framing effects did not vary with participants' prior familiarity: neither chatbot use frequency nor self-reported comfort with AI moderated framing's effect on pre-interaction impressions or on impression change (Appendix Table~\ref{tab:appendix-familiarity}).

\subsection{Framing changes interaction behavior}
\label{sec:rq2}

\begin{figure}[t]
  \centering
  \includegraphics[width=0.9\linewidth]{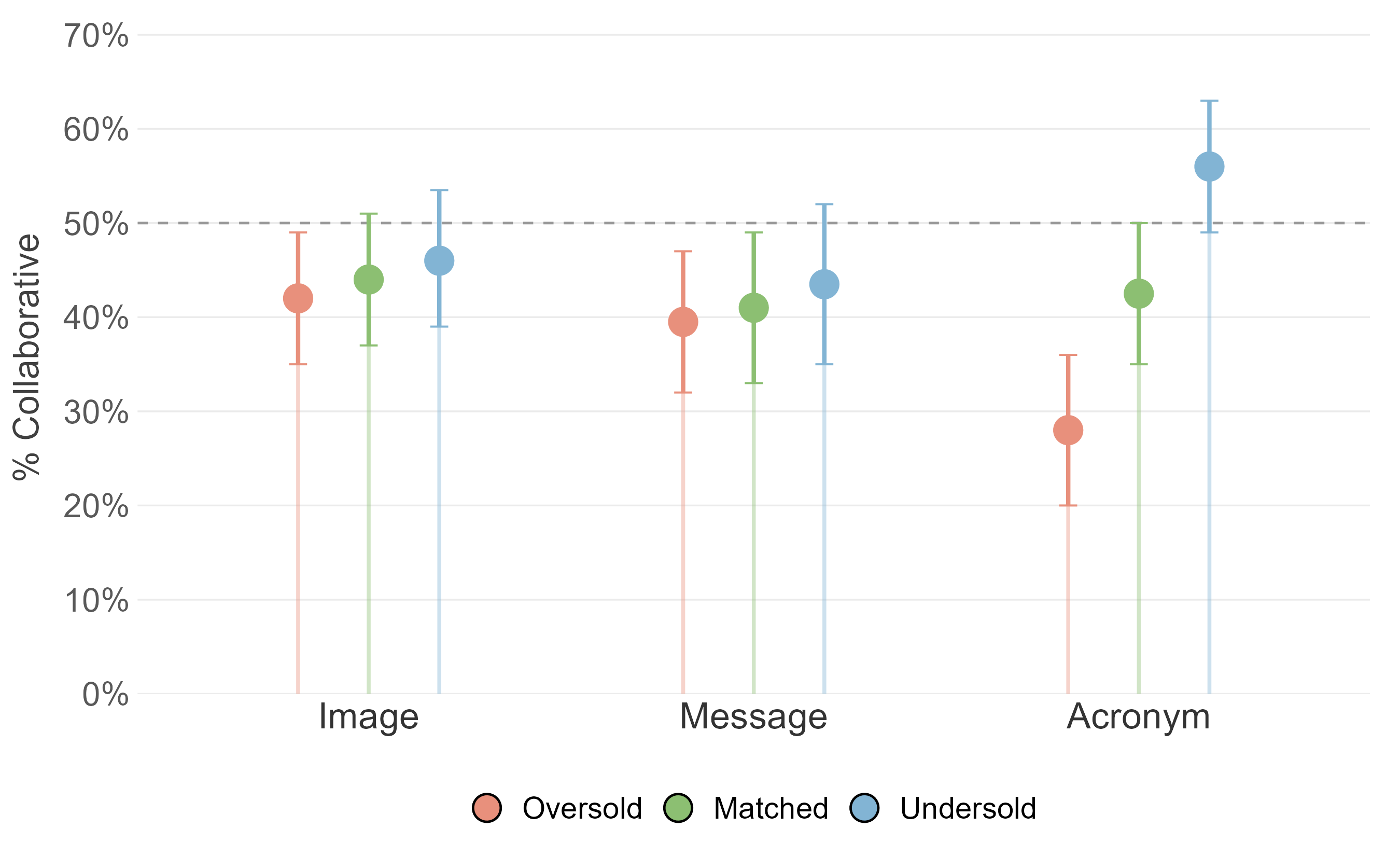}
    \vspace{-2mm}
  \caption{Percentage of user messages classified as Collaborative (vs.\ Directive) by task and framing condition. Error bars are $95\%$ CIs. Dashed line at $50\%$ marks an equally directive/collaborative split.}
  \label{fig:rq2-collab}
\end{figure}

Framing also changed how participants worked with the model. On the acronym task, Oversold participants sent the most messages, with the shortest gaps and shortest prompts, consistent with re-prompting in search of an output the model was failing to deliver; Undersold participants sent fewer, longer messages with more deliberation between them, consistent with active co-construction; Matched fell between the two on every metric (Figure~\ref{fig:rq2-engagement}). Read in task order, the gradient is weak and inconsistent on image generation (Appendix Figure~\ref{fig:appendix-img-bars}), attenuated on outreach message writing (Appendix Figure~\ref{fig:appendix-msg-bars}), and steepest on acronym building: the divergence grows across the session, and only the acronym task left user input as the primary determinant of a good output.

The same split appears in what participants wrote and in what they submitted. On the acronym task, Undersold participants sent markedly more collaborative messages than Oversold, with Matched in between, while the proportion varied little with framing on the other two tasks (Figure~\ref{fig:rq2-collab}). Oversold participants' acronym submissions were also roughly twice as close to the model's own wording as Undersold participants', measured as normalized Levenshtein similarity to the most similar assistant message in their log ($0.32$ against $0.17$, $p{=}.001$; Appendix Table~\ref{tab:appendix-reliance}); on outreach the measure was near ceiling across conditions, since the model's draft already met the brief's requirements. Representative chat logs are in Appendix Figures~\ref{fig:appendix-chatlog-Oversold} and~\ref{fig:appendix-chatlog-Undersold}.

Participants' own reports after each task follow the same order: the framing gap in self-efficacy change and in expectations met is small on image generation and outreach, then widens on acronym building, where the difference in expectations met reaches significance (Appendix Table~\ref{tab:appendix-pertask}).

\begin{figure}[t]
\vspace{-3mm}
  \centering
  \includegraphics[width=0.9\linewidth]{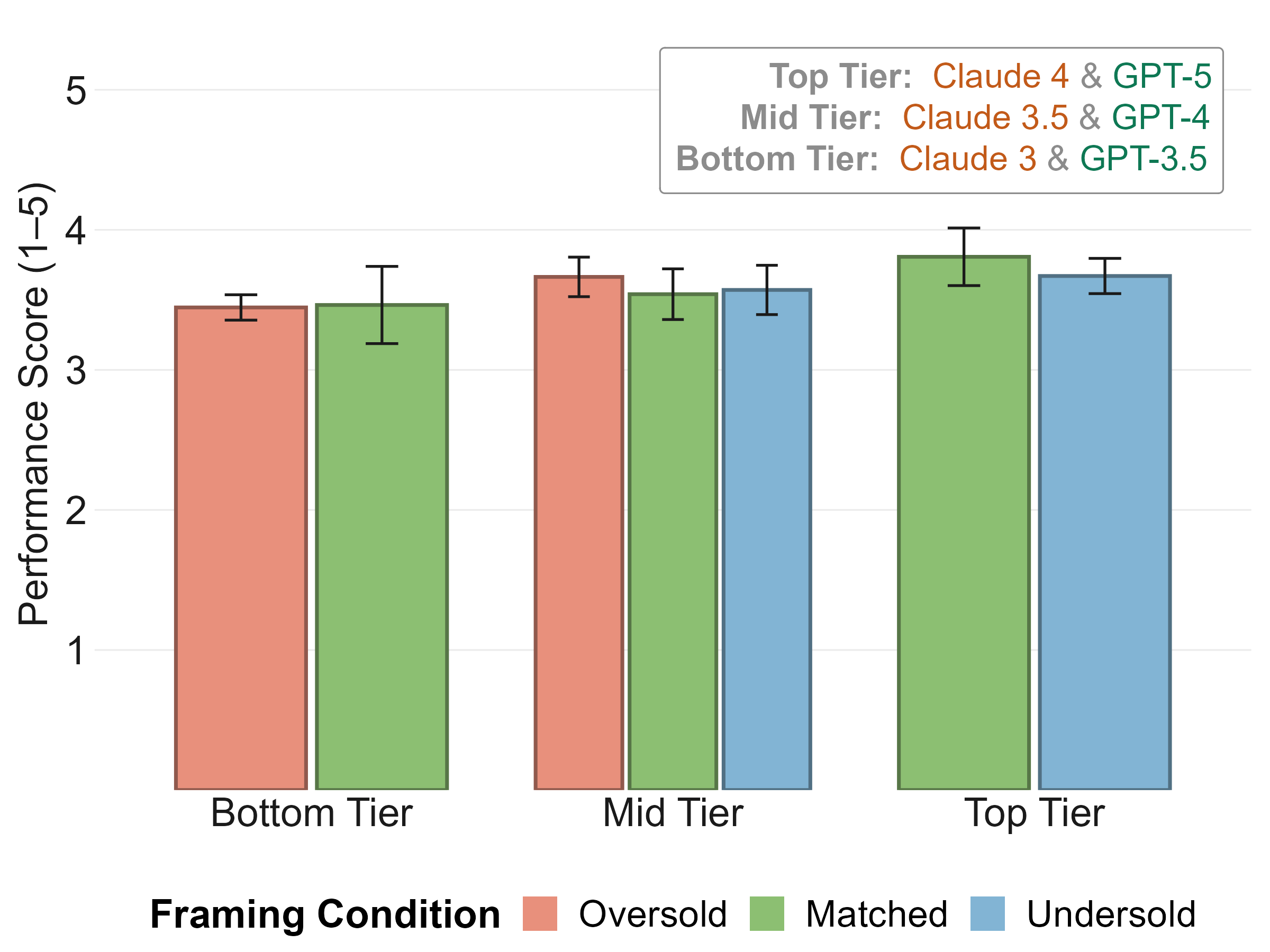}
  \vspace{-1mm}
  \caption{Mean task performance by source model tier and framing condition, with $95\%$ CIs. There are seven bars rather than nine because a bottom-tier model cannot be Undersold and a top-tier model cannot be Oversold, as no lower or higher tier exists to show (Table~\ref{tab:design}).}
  \label{fig:rq3-perf}
\end{figure}

\subsection{Output quality corresponds to model capability, not framing}
\label{sec:rq3}

Framing did not affect the quality of what users and the model produced together. Performance increased linearly with model tier and was flat across framing conditions within each tier (Figure~\ref{fig:rq3-perf}). Oversold participants did not produce better outputs because they expected a better model and Undersold participants did not produce worse ones because they expected a weaker one.

\begin{figure}[t]
  \centering
  \includegraphics[width=0.9\linewidth]{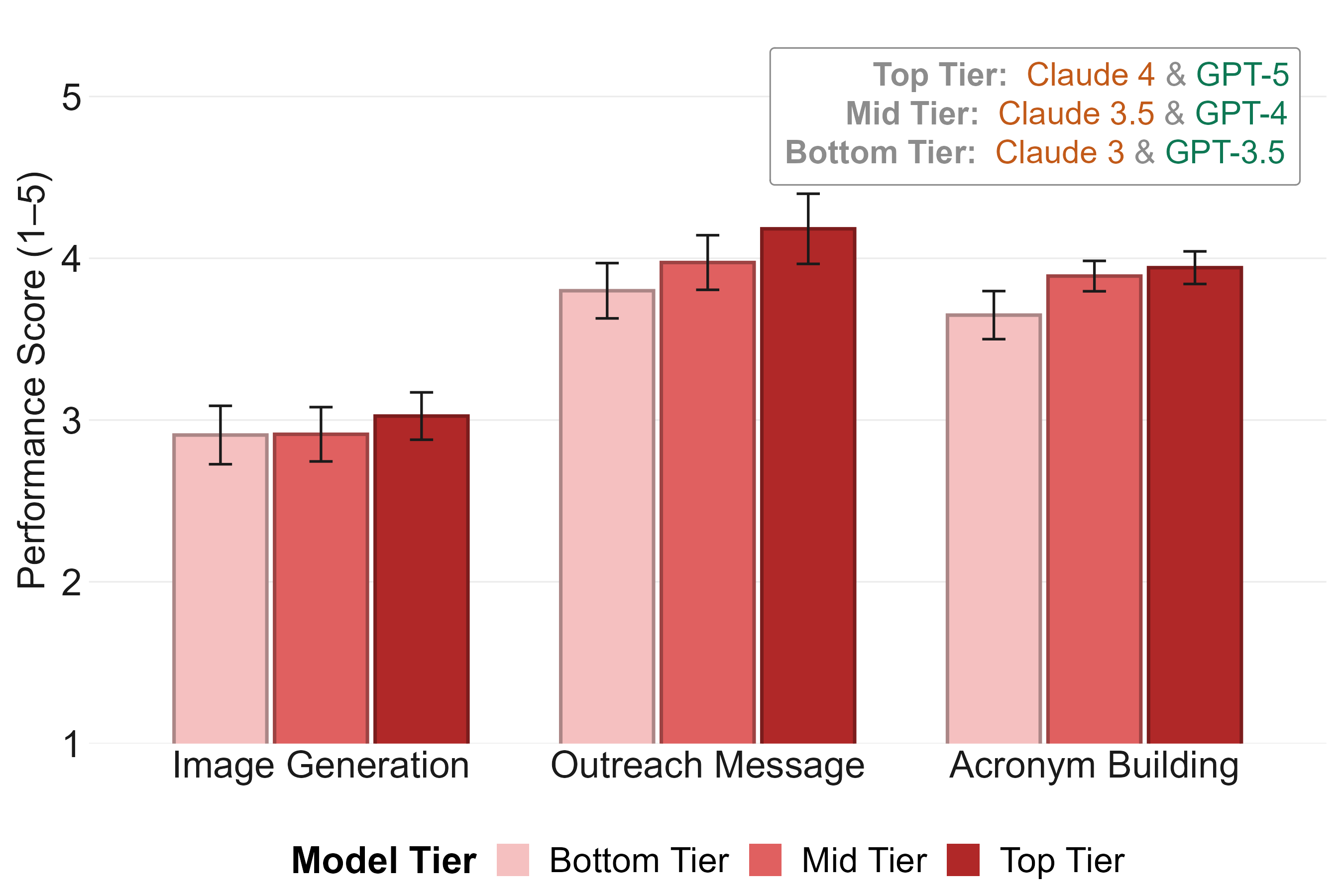}
  \vspace{-4mm}
  \caption{Mean task performance by task and source model tier, with $95\%$ CIs.}
  \label{fig:rq3-task-tier}
\end{figure}

A regression of z-scored performance on framing and model tier confirms this. Model tier predicted performance ($\beta{=}0.44$, $p{=}0.0006$) and framing did not ($\beta{=}0.14$, $p{=}0.276$). A two-way ANOVA treating both as categorical factors gives the same separation without assuming the tier effect is linear (tier $F{=}6.12$, $p{=}0.003$, $\eta^2_p{=}0.072$; framing $F{=}0.68$, $p{=}0.51$; Appendix Table~\ref{tab:appendix-anova}). The same pipeline therefore registers real capability differences while registering none across framing. The model tier gradient appeared on two of the three tasks (Figure~\ref{fig:rq3-task-tier}), reaching significance on outreach message writing and acronym building; on image generation, where every image was rendered by the same backend model, scores clustered tightly and neither factor separated the groups.

%The judge registered gaps between model tiers, confirming it could detect quality differences, but found none between framing conditions.

Self-efficacy change and performance were weakly correlated ($r{=}0.21$, $p{=}0.008$; Appendix Figure~\ref{fig:appendix-se-perf}): participants who scored higher gained slightly more self-efficacy, but the relationship was loose and the framing conditions overlapped heavily on performance. How capable participants felt and how well they actually performed therefore tracked each other only loosely.

\subsection{Impression change is driven by experience, not output}
\label{sec:rq4}

% Framing shifts impressions, but not output quality. What, then, drives their change in impressions? 

We ran bivariate OLS regressions of $\Delta_{\mathrm{UTAUT}}$ and $\Delta_{\mathrm{Godspeed}}$ on three candidate predictors: objective task performance, self-efficacy change ($\Delta_{\mathrm{SE}}$), and expectations met. All predictors and outcomes were z-scored, so coefficients are standardized effect sizes. Results are in Table~\ref{tab:rq4-ols}.

\begin{figure}[t]
    \centering
    \vspace{-3mm}
    \includegraphics[width=0.9\linewidth]{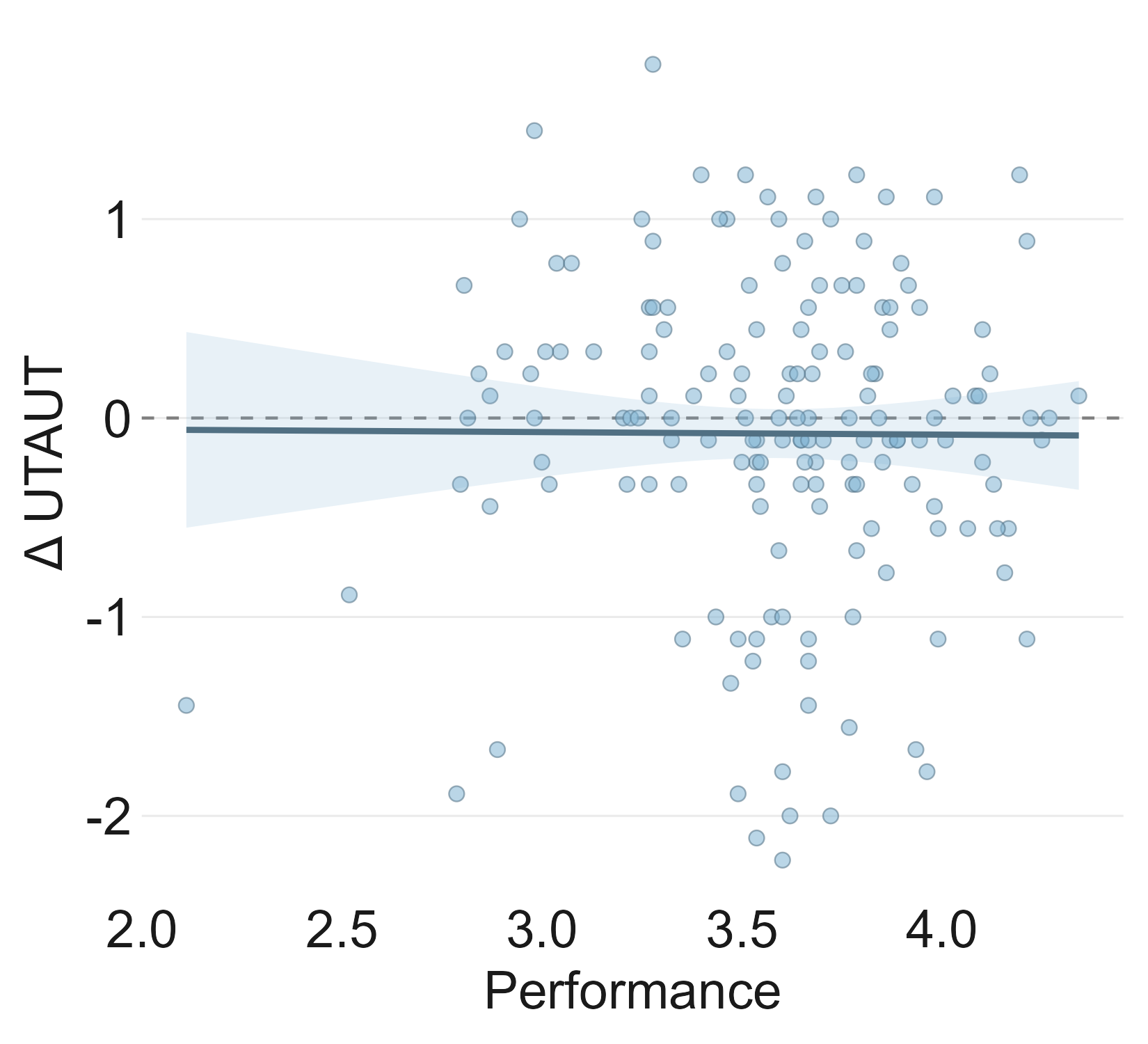}
    \caption{UTAUT change as a function of overall performance. OLS fit; $95\%$ CI band, $N{=}162$.}
    \vspace{-2mm}
    \label{fig:rq4-perf-utaut}
\end{figure}

How well participants performed did not predict how their impression changed. The effect on $\Delta_{\mathrm{UTAUT}}$ was essentially zero ($\beta{=}-0.01$, $p{=}0.938$; Figure~\ref{fig:rq4-perf-utaut}), and the effect on $\Delta_{\mathrm{Godspeed}}$, though larger, was still not significant ($\beta{=}0.11$, $p{=}0.180$; Appendix Figure~\ref{fig:appendix-rq4-perf-gs}).

% Even though output quality differed substantially across model tiers, what participants produced had no bearing on how their impression of the model shifted.

\begin{figure}[t]
    \centering
    \includegraphics[width=0.83\linewidth]{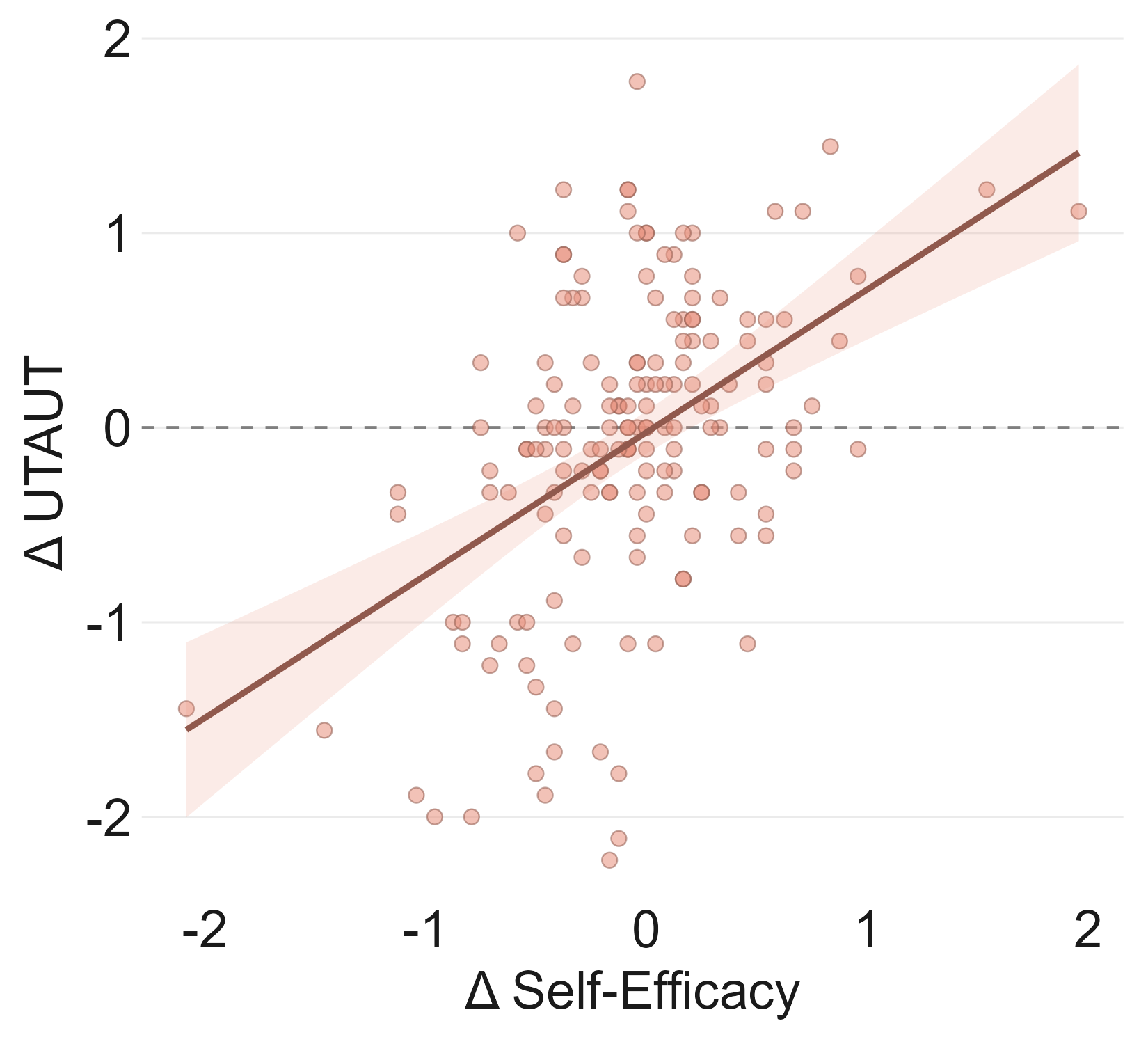}
    \caption{UTAUT change as a function of overall self-efficacy change ($\Delta_{\mathrm{SE}}$). OLS fit; $95\%$ CI band, $N{=}162$.}
    \label{fig:rq4-se-utaut}
      \vspace{-2mm}
\end{figure}

Self-efficacy change was a strong predictor: participants who finished the study feeling more capable as collaborators rated the model more favorably, and a one standard-deviation gain in self-efficacy corresponded to a 0.47 standard-deviation rise in $\Delta_{\mathrm{UTAUT}}$ ($p < .001$; Figure~\ref{fig:rq4-se-utaut}) and a 0.36 standard-deviation rise in $\Delta_{\mathrm{Godspeed}}$ ($p < .001$; Appendix Figure~\ref{fig:appendix-rq4-se-gs}). This aligns with evidence that active AI collaboration preserves users' sense of competence better than passive reliance does \citep{lee2026relying}.

Whether the model met participants' expectations had a comparably strong effect on impression change. Those who felt the model fell short revised their impression downward; those who felt it exceeded revised upward. A one standard-deviation increase in expectations met corresponded to a 0.47 standard-deviation gain in $\Delta_{\mathrm{UTAUT}}$ and a 0.50 standard-deviation gain in $\Delta_{\mathrm{Godspeed}}$ (both $p < .001$; Appendix Figures~\ref{fig:appendix-rq4-exp-utaut} and~\ref{fig:appendix-rq4-exp-gs}).

Expectations met on each of the three tasks predicted impression change, most strongly on the final acronym task and least on outreach, whose ratings were near-uniform across conditions and so left less variance to explain (Appendix Table~\ref{tab:appendix-pertask-exp}).

Both experiential predictors tracked impression change, while objective performance did not. What moves users' ratings of a model is the experience of the interaction, not the output it produced.

\section{Related Work}
\label{sec:related}

\paragraph{Framing Effects in Human-AI Interaction}
The presentation of AI systems shapes user perception and behavior independently of actual capability \cite{10.1145/3529225, GRIMES2021113515}. Pre-interaction framing significantly alters baseline acceptance and judgments of system trustworthiness \cite{10.1145/3290605.3300641, pataranutaporn2023influencing}, and these expectation cues bias evaluations upward or downward independently of actual quality \cite{10.1145/3772318.3790492}, influence reliance decisions \cite{10.1145/3579612}, and shape active use as users develop implicit assumptions about underlying capabilities \cite{10.1145/3706598.3713886, 10.1145/3706598.3713751}. Beliefs about capability drive reliance from both sides of the interaction: a model's stated accuracy shapes user trust alongside the accuracy users observe in practice \cite{10.1145/3290605.3300509}, while overestimating one's own competence reduces appropriate reliance on the system \cite{10.1145/3544548.3581025}. Our work builds on this by controlling the underlying model to create explicit expectation violations and examining how capability framing shifts user impressions of their assigned model.

\paragraph{Subjective versus Objective LLM Evaluation}
Standard LLM evaluations measure true capability through automated, model-focused benchmarks \cite{kazemi-etal-2025-big, zhong-etal-2024-agieval, hendryckstest2021, liang2023holistic}, yet these metrics often fail to capture the contextual nuances of how users evaluate models in practice \cite{11002710, 11488866}. Human-centric frameworks instead capture subjective experiences through in-the-moment satisfaction \cite{liu-etal-2025-understand} and collaborative self-efficacy \cite{ju2025developing}, but subjective perceptions frequently diverge from objective reality: evaluators rely on flawed linguistic heuristics \cite{doi:10.1073/pnas.2208839120}, miscalibrate confidence \cite{steyvers2025large}, and rate AI-labeled content lower regardless of quality \cite{zhu-etal-2025-human}. Contrasting subjective self-efficacy with objective output quality allows us to directly assess which more strongly drives changes in user impressions.

\begin{table}[t]
  \centering
  \small
  \begin{tabular}{llcc}
    \toprule
    \textbf{Predictor} & \textbf{Outcome} & \boldmath{$\beta$} & \boldmath{$p$} \\
    \midrule
    \multirow{2}{*}{Performance}            & $\Delta$ UTAUT    & $-$0.01          & 0.938 \\
    & $\Delta$ Godspeed & 0.11             & 0.180 \\
    \addlinespace[2pt]
    \multirow{2}{*}{$\Delta$ Self-Efficacy} & $\Delta$ UTAUT    & \textbf{0.47}*** & $<\!.001$ \\
    & $\Delta$ Godspeed & \textbf{0.36}*** & $<\!.001$ \\
    \addlinespace[2pt]
    \multirow{2}{*}{Expectations met}       & $\Delta$ UTAUT    & \textbf{0.47}*** & $<\!.001$ \\
    & $\Delta$ Godspeed & \textbf{0.50}*** & $<\!.001$ \\
    \bottomrule
  \end{tabular}
  \vspace{-1mm}
  \caption{Bivariate OLS regressions of impression change on each candidate predictor ($N{=}162$ per regression). Predictors and outcomes are independently z-scored, so $\beta$ is fully standardized. Asterisks on $\beta$: * $p < 0.05$, ** $p < 0.01$, *** $p < 0.001$.}
  \label{tab:rq4-ols}
\end{table}

\paragraph{Interaction Behavior with AI Models}
Human interaction with LLMs ranges from directive commands to collaborative partnering \cite{10.1007/978-3-031-60615-1_5, thu2025prompting}. Researchers typically operationalize these dynamics through structural metrics like query adjustments and turn patterns \cite{10.1145/3726302.3729998, 10.1145/3627508.3638344, koyuturk2025understanding}, which are shaped more by a system's presentation than its underlying architecture \cite{10.1093/iwc/iwag017}. Treating capability framing as the manipulated variable lets us observe how prompting behavior shifts while output quality stays fixed.

\section{Discussion and Conclusion}
\label{sec:discussion}
Framing effects appear in our data where prior work predicts them, in attitudes and in reliance, and go no further: the quality of the artifact is unaffected. The difference is in what the task asks of the user. In advice-taking studies the participant decides whether to accept the system's recommendation, and that decision is both a measure of trust and the thing that determines whether the answer is correct \citep{10.1145/3290605.3300509, 10.1145/3544548.3581025}, so an effect on trust becomes an effect on performance almost by definition. In sustained collaboration the model produces the output no matter how the user prompts it, so trust and result can move independently.

A user's evaluation of an LLM is not a clean readout of its capability. For individual users, this means head-to-head self-reports between models of unequal reputation should be read with care: such reports encode the model's reputation alongside its actual performance \citep{10.1145/3772318.3790492, pfeuffer2026impact}. For organizations deploying AI internally, the implication is more actionable: how a tool is introduced may matter as much as which tool was chosen. Overselling to drive adoption risks depressing post-deployment satisfaction; setting realistic expectations may encourage more collaborative engagement with the tool \citep{10.2307/249008, ubellacker2025making}.

Because impression change after use is driven by whether expectations were met rather than by what users produced, the way capability is communicated to users matters distinctly from the capability itself. For developers and benchmark designers, leaderboard rankings function as expectation-setting commitments \citep{bean2026measuring}, but the capability differences these rankings advertise are less perceptible to users during use than the rankings imply. A model marketed honestly is more likely to land in users' \textit{exceeded expectations} bucket than one marketed at its benchmark ceiling. 

For evaluation research, user-preference data that drives pipelines like Chatbot Arena \citep{10.5555/3692070.3692401} partly measures expectation management \citep{wu2025style, ICLR2024_f719b8ca}. Without controls that hold expectations constant across compared models, such data should be read as \textit{experience relative to expectation}, not capability in isolation. Any evaluation routed through user judgment therefore needs identity blinding or a measured-expectation covariate, whether it is preference data, an in-product rating, or a satisfaction survey. Artifacts co-produced with users are more robust: participants' priors threaten the ratings attached to those artifacts, not the work itself.

LLM capability is real, and benchmarks measure something. But what users walk away believing is shaped at least as much by what they were told about a model as by what it did. The model and the message about it are not separate things to a user; they arrive together, and they leave together.

\paragraph{Future work.}
The larger question is how user-facing evaluation can be built to survive the priors users bring to it. Expectation goes unmeasured in the pipelines that collect user judgments today, and a short instrument administered alongside a rating would let it be reported or controlled for rather than absorbed into the score. What blinding buys is also open, since an anonymous protocol removes a model's identity only when users cannot recover it from the output. The same holds for automated evaluation, where it remains less clear whether naming the model behind an output shifts a judge's scores as our landing page shifted participants'. And because our effects concentrated on the most open-ended task, evaluation suites could be triaged by how much room each component leaves for expectation to work. Extensions of the study itself are noted in Limitations.

\section*{Limitations}
\label{sec:limit}

\paragraph{Population.} Our sample was US-based, English-speaking, and drawn from Prolific with a high prior approval rate. Crowdsourced workers may differ from casual consumer users or workplace deployments in ways that could amplify or attenuate framing effects; replication across more naturalistic samples would test the generalizability of these effects. As well, sensitivity to framing may also vary with cultural context, which a cross-locale replication would test.

\paragraph{Sample sizing.} The $N{=}162$ target was derived from a Monte Carlo power analysis seeded with a pilot ($n{=}18$). However, effect estimates from $n{=}18$ are themselves noisy; the final sample is therefore sized against a deliberately conservative effect estimate rather than a precise one.

\paragraph{Models and tier ordering.} Tier labels were anchored to public benchmark rankings at the time of data collection. The absolute capability gap between adjacent tiers is not equal across families, and the two bottom-tier models are distinct systems with different stylistic tendencies. We used two model families specifically to test the manipulation's robustness across different lineages; the families differ modestly in overall score (Appendix Table~\ref{tab:appendix-family}), and the central results are unchanged with family as a fixed effect. A finer-grained mapping between perceived capability and benchmark position would strengthen future replications. Our background measures also capture general familiarity with chatbots rather than preconceptions about the specific models named on the landing page, so model-specific reputations remain untested.

The manipulation also depends on model names whose tier ordering a general participant pool already recognizes, which is why we used the GPT and Claude lineages. Open-weight families would test whether framing effects survive where that ordering is less publicly established.

\paragraph{Task order.} Tasks were presented in a fixed order: image generation, outreach message, then acronym building. The acronym task is also the most open-ended of the three, so the larger framing effect observed there is consistent with two non-mutually-exclusive accounts: (a) framing effects concentrate on tasks where user input most shapes what counts as a good output, and (b) framing effects compound over session time as users accumulate experience filtered through their initial expectation. Our design cannot distinguish these. A counterbalanced replication would separate task openness from time-in-session and would also test whether framing effects strengthen, attenuate, or reverse across repeated exposures within a session.

\paragraph{LLM judges and family-specific agreement.} Task performance was rubric-scored by GPT-5 and message style was classified by \texttt{claude-haiku-4-5}. We averaged three independent scoring runs per output and validated the judge against three independent human raters on a subsample of submissions (Appendix~\ref{sec:appendix-validation}). Agreement among the human raters was modest, reflecting the inherent subjectivity of the rubrics, particularly for humor and persuasiveness. Computed on the same statistic, the judge agreed with the raters about as closely as the raters agreed with one another ($\rho{=}0.301$ and $0.298$). Agreement was notably stronger for Claude-family outputs ($\rho{=}0.483$) than for GPT-family outputs ($\rho{=}0.179$, n.s.). Our performance results in \S\ref{sec:rq3} and in \S\ref{sec:rq4} should therefore be read as more reliable on the Claude side, although the central dissociation (framing affects behavior and impressions but not output) does not depend on the absolute calibration of the judge.

Similarly, the message classifier, run three times per message with a majority vote, was validated in the same way against three human annotators on a stratified sample of 100 messages (Appendix~\ref{sec:appendix-classifier-validation}). Agreement among the annotators was moderate (Fleiss $\kappa{=}0.506$), reflecting how interpretive the collaborative/directive distinction is at the message level, and the classifier reached fair agreement against that ceiling (Cohen $\kappa{=}0.269$). Because the classifier was blind to framing condition, its errors are distributed identically across groups and attenuate group differences rather than creating them, so the collaboration gradient in \S\ref{sec:rq2} persisted despite this noise rather than because of it.

\paragraph{Cross-sectional design.} Impression change was measured within a single session, which is appropriate for studying how framing shapes an initial impression but does not address how those effects evolve with repeated exposure. Whether framing-induced impressions persist, erode, or compound over time is a question for longitudinal designs.

\paragraph{Statistical testing.} We report uncorrected $p$-values. Our safeguard against false positives is built into the design rather than into a correction: every effect of interest replicates across the two independent impression measures (UTAUT and Godspeed), and in \S\ref{sec:rq4} across two independent experiential predictors (self-efficacy change and expectations met). This cross-measure consistency provides design-level robustness for the central effects.

\paragraph{Predictor independence.} Expectations-met and impression change are related but distinct: the former is rated per-task on a three-point scale immediately after each task; the latter is the difference between session-level composites on two multi-item validated scales. The two are measured on different scales, at different time points, and against different referents, so the strong association between them is informative rather than mechanical. Self-efficacy change, measured against the user's own capability rather than the model's, provides a fully independent experiential predictor; that both track impression change while objective performance does not is the pattern of interest.

\section*{Ethics Statement}
\label{sec:ethics}

All study procedures were approved by the authors' institutional research ethics board prior to data collection. Participants were recruited through Prolific and provided informed consent before any study activities. They were compensated at a rate of \$12.50 USD per hour to be consistent with Prolific's recommended fair-pay guidelines, with an additional \$10 performance bonus available on each of the tasks. No personally identifying information was collected; participant records were keyed only by Prolific worker ID, which was discarded after compensation processing. Conversation logs and survey responses were stored on access-controlled infrastructure. Participants were informed in the consent form that they would be interacting with a large language model and that their interactions would be analyzed; the specific framing manipulation was disclosed in a debrief at the end of the study.

The framing manipulation involved presenting some participants with a model description that did not match the model they actually used. This deception was minimal, time-limited, fully reversed in the debrief, and judged by the ethics review board to pose no risk of harm. We see no reasonably foreseeable harms from the methods or findings of this work; if anything, the findings argue for more transparent communication of AI system capability to users.

\bibliography{custom}

\clearpage
\appendix
\section{Appendix}
\label{sec:appendix}

This appendix collects the survey instruments, task materials, evaluation rubrics, example outputs, and supplementary analyses referenced in the main paper.

\subsection{Survey Instruments}
\label{sec:appendix-surveys}

This section reproduces the items used in each scale, describes how each was administered, and describes how composite scores were computed.

\paragraph{Administration.}The UTAUT performance expectancy and effort expectancy subscales, along with one additional engagement item, were administered in future tense before the study and in past tense after the study, with one embedded attention check per administration (Table~\ref{tab:appendix-utaut}). The Godspeed Perceived Intelligence subscale (Table~\ref{tab:appendix-godspeed}) used the same five bipolar adjective items both pre- and post-study, as well as one additional bipolar adjective of ``Artificial vs Natural''. The task-specific NGSE items (Table~\ref{tab:appendix-ngse}) were administered before each of the three tasks in future tense and after each task in past tense.

\paragraph{UTAUT composites.} Each administration yields a single score: the arithmetic mean of all non-attention items on a 1--5 Likert scale. Pre-study UTAUT is the mean across the pre-study items (Table~\ref{tab:appendix-utaut}, left column); post-study UTAUT is the mean across the post-study items (right column). $\Delta_{\mathrm{UTAUT}}$ is post-study minus pre-study.

\paragraph{Godspeed Perceived Intelligence composites.} The six bipolar items (Table~\ref{tab:appendix-godspeed}) are each rated on a 1--7 semantic differential scale and averaged. The same items appear pre- and post-study; $\Delta_{\mathrm{Godspeed}}$ is post-study minus pre-study.

\paragraph{Task-specific NGSE composites.} Each task has eight task-specific items (Table~\ref{tab:appendix-ngse}), each rated on a 1--5 Likert scale. The pre-task and post-task scores for a given task are the mean across non-attention items. Per-task self-efficacy change is post-task minus pre-task. Overall self-efficacy change ($\Delta_{\mathrm{SE}}$) is the unweighted mean of the three per-task changes.

\paragraph{Expectations met.} After each task, participants chose one of three options describing whether the model fell short of, met, or exceeded their expectations for that task. The three options were coded as $1$, $2$, and $3$ respectively. The composite \textit{expectations-met} score used in \S\ref{sec:rq4} is the average of the three per-task codings, ranging from $1$ (fell short on all three) to $3$ (exceeded on all three).

\subsection{Participant Exclusion Criteria}
\label{sec:appendix-filtering}

Every conversation log was reviewed manually after data collection and before any outcome analysis, blind to framing condition.

\paragraph{Insufficient engagement.} Participants were told that each task required at least five minutes of interaction. A participant was removed if at least two of their three logs showed conversations substantially shorter than five minutes, or containing only one or two prompts. Participants who fell short on a single task but engaged substantially on the other two were retained.

\paragraph{Non-adherence.} Non-adherence meant conversations wholly unrelated to the task, such as asking the model unrelated trivia questions, generating images unrelated to the target, or inventing new letter sets on the acronym task.

\paragraph{Counts.} Ten participants were removed, six for non-adherence and four for insufficient engagement. Each was replaced through new recruitment, preserving nine participants per source-by-shown cell and the balanced $N{=}162$.

\begin{table*}[!htbp]
  \centering
  \small
  \begin{tabular}{p{0.46\linewidth} p{0.46\linewidth}}
    \toprule
    \textbf{Pre-study (future tense)} & \textbf{Post-study (past tense)} \\
    \midrule
    \multicolumn{2}{l}{\textbf{Performance Expectancy}} \\
    \midrule
    1. I think that I will find this chatbot useful & 1. I found this chatbot useful \\
    2. I expect that using this chatbot will enable me to accomplish tasks more quickly & 2. Using this chatbot enabled me to accomplish tasks more quickly \\
    3. I expect that using this chatbot will increase my productivity & 3. Using this chatbot increased my productivity \\
    4. If I use this chatbot, I feel I will increase my chances of getting my work done successfully & 4. Using this chatbot, I felt my chances of getting my work done successfully was increased \\
    \midrule
    \multicolumn{2}{l}{\textbf{Effort Expectancy}} \\
    \midrule
    5. I think I will find this chatbot easy to use & \cellcolor{cellred}5. Please select ``Disagree'' for this item. \\
    6. I think my interactions with this chatbot will be clear and understandable & 6. I found this chatbot easy to use \\
    7. I feel that it will be easy for me to become skillful at using this chatbot & 7. My interactions with this chatbot were clear and understandable \\
    \cellcolor{cellred}8. Select ``Strongly Agree'' for this statement. & 8. It was easy for me to become skillful at using this chatbot \\
    9. I expect that learning to operate this chatbot will be easy for me & 9. Learning to operate this chatbot was easy for me \\
    \midrule
    \multicolumn{2}{l}{\textbf{Additional Engagement Item}} \\
    \midrule
    10. I plan to use the chatbot as much as possible during the tasks & 10. I used the chatbot as much as possible during the tasks \\
    \bottomrule
  \end{tabular}
  \caption{UTAUT Performance Expectancy and Effort Expectancy items with an additional engagement item added to the composite, each rated on a 1--5 Likert scale. Cells highlighted in \textit{red} are attention checks; these were excluded from all subscale composites.}
  \label{tab:appendix-utaut}
\end{table*}

\begin{table*}[!htbp]
  \centering
  \small
  \begin{tabular}{p{0.48\linewidth} p{0.48\linewidth}}
    \toprule
    \textbf{Pre-task (future tense)} & \textbf{Post-task (past tense)} \\
    \midrule
    \multicolumn{2}{l}{\textbf{Image generation task}} \\
    \midrule
    1. I will be able to generate an image closest to the target image using the chatbot. & 1. I was able to generate an image closest to the target image using the chatbot. \\
    2. When facing difficulty with the image generation task, I am certain I will still accomplish it with the chatbot. & 2. When facing difficulty with the image generation task, I was certain I would still accomplish it with the chatbot. \\
    3. In general, I think that I can obtain outcomes that are important to achieving the target image using the chatbot. & 3. In general, I thought that I obtained outcomes that were important to achieving the target image using the chatbot. \\
    4. I believe I can succeed at the image generation task using the chatbot. & 4. I believe I succeeded at the image generation task using the chatbot. \\
    5. I will be able to successfully overcome challenges in the image generation task using the chatbot. & 5. I was able to successfully overcome challenges in the image generation task using the chatbot. \\
    6. Using the chatbot, I am confident that I can perform the image generation task effectively. & 6. Using the chatbot, I was confident that I performed the image generation task effectively. \\
    7. Compared to other people, I can do the image generation task very well with the chatbot. & 7. Compared to other people, I think I did the image generation task very well with the chatbot. \\
    8. Even when generating the target image is tough, I can perform quite well with the chatbot. & 8. Even when generating the target image was tough, I performed quite well with the chatbot. \\
    \midrule
    \multicolumn{2}{l}{\textbf{Outreach message writing task}} \\
    \midrule
    1. I will be able to write the most convincing message using the chatbot. & 1. I was able to write the most convincing message using the chatbot. \\
    2. When facing difficulty with the message writing task, I am certain I will still accomplish it with the chatbot. & 2. When facing difficulty with the message writing task, I was certain I would still accomplish it with the chatbot. \\
    3. In general, I think that I can obtain outcomes that are important to writing a convincing message using the chatbot. & 3. In general, I thought that I obtained outcomes that were important to writing a convincing message using the chatbot. \\
    \cellcolor{cellred}4. Even if you don't agree, select Neutral below. & 4. I believe I succeeded at the message writing task using the chatbot. \\
    5. I believe I can succeed at the message writing task using the chatbot. & 5. I was able to successfully overcome challenges in the message writing task using the chatbot. \\
    6. I will be able to successfully overcome challenges in the message writing task using the chatbot. & 6. Using the chatbot, I was confident that I performed the message writing task effectively. \\
    7. Using the chatbot, I am confident that I can perform the message writing task effectively. & 7. Compared to other people, I think I did the message writing task very well with the chatbot. \\
    8. Compared to other people, I can do the message writing task very well with the chatbot. & 8. Even when writing the message was tough, I performed quite well with the chatbot. \\
    9. Even when writing the message is tough, I can perform quite well with the chatbot. & \\
    \midrule
    \multicolumn{2}{l}{\textbf{Acronym building task}} \\
    \midrule
    1. I will be able to create the funniest acronyms using the chatbot. & 1. I was able to create the funniest acronyms using the chatbot. \\
    2. When facing difficulty with the acronym creation task, I am certain I will still accomplish it with the chatbot. & 2. When facing difficulty with the acronym creation task, I was certain I would still accomplish it with the chatbot. \\
    3. In general, I think that I can obtain outcomes that are important to creating a funny acronym using the chatbot. & 3. In general, I thought that I obtained outcomes that were important to creating a funny acronym using the chatbot. \\
    4. I believe I can succeed at the acronym creation task using the chatbot. & 4. I believe I succeeded at the acronym creation task using the chatbot. \\
    5. I will be able to successfully overcome challenges in the acronym creation task using the chatbot. & 5. I was able to successfully overcome challenges in the acronym creation task using the chatbot. \\
    6. Using the chatbot, I am confident that I can perform the acronym creation task effectively. & 6. Using the chatbot, I was confident that I performed the acronym creation task effectively. \\
    7. Compared to other people, I can do the acronym creation task very well with the chatbot. & 7. Compared to other people, I think I did the acronym creation task very well with the chatbot. \\
    8. Even when the acronym creation is tough, I can perform quite well with the chatbot. & 8. Even when the acronym creation was tough, I performed quite well with the chatbot. \\
    \bottomrule
  \end{tabular}
  \caption{Task-specific NGSE items for the three study tasks, rated on a 1--5 Likert scale. Cells highlighted in \textit{red} are attention checks; these were excluded from all subscale composites.}
  \label{tab:appendix-ngse}
\end{table*}

\begin{table}[H]
  \centering
  \begin{tabular}{l}
    \toprule
    \textbf{Perceived Intelligence} \\
    \midrule
    Incompetent -- Competent \\
    Ignorant -- Knowledgeable \\
    Irresponsible -- Responsible \\
    Unintelligent -- Intelligent \\
    Foolish -- Sensible \\
    \toprule
    \textbf{Additional Pair} \\
    \midrule
    Artificial -- Natural \\
    \bottomrule
  \end{tabular}
  \caption{Godspeed Perceived Intelligence items and additional bipolar pair, each rated on a semantic differential scale from 1--7, identical across pre- and post-study administrations.}
  \label{tab:appendix-godspeed}
\end{table}

\begin{figure*}[!htbp]
  \centering
  \includegraphics[width=\linewidth]{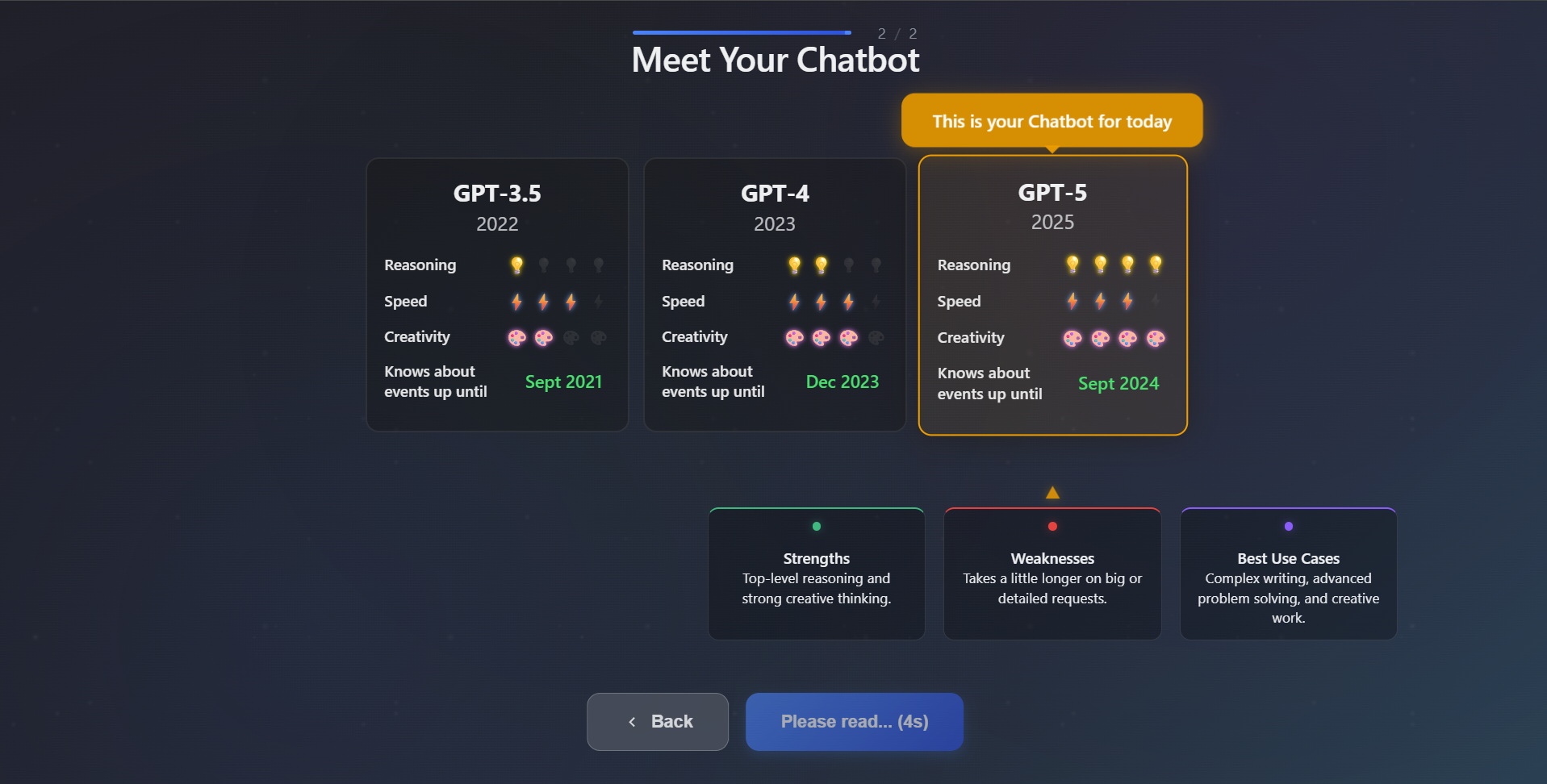}
  \caption{The landing page that constituted the capability framing manipulation.}
  \label{fig:appendix-framing}
\end{figure*}

\begin{figure*}[!htbp]
  \centering
  \includegraphics[width=\linewidth]{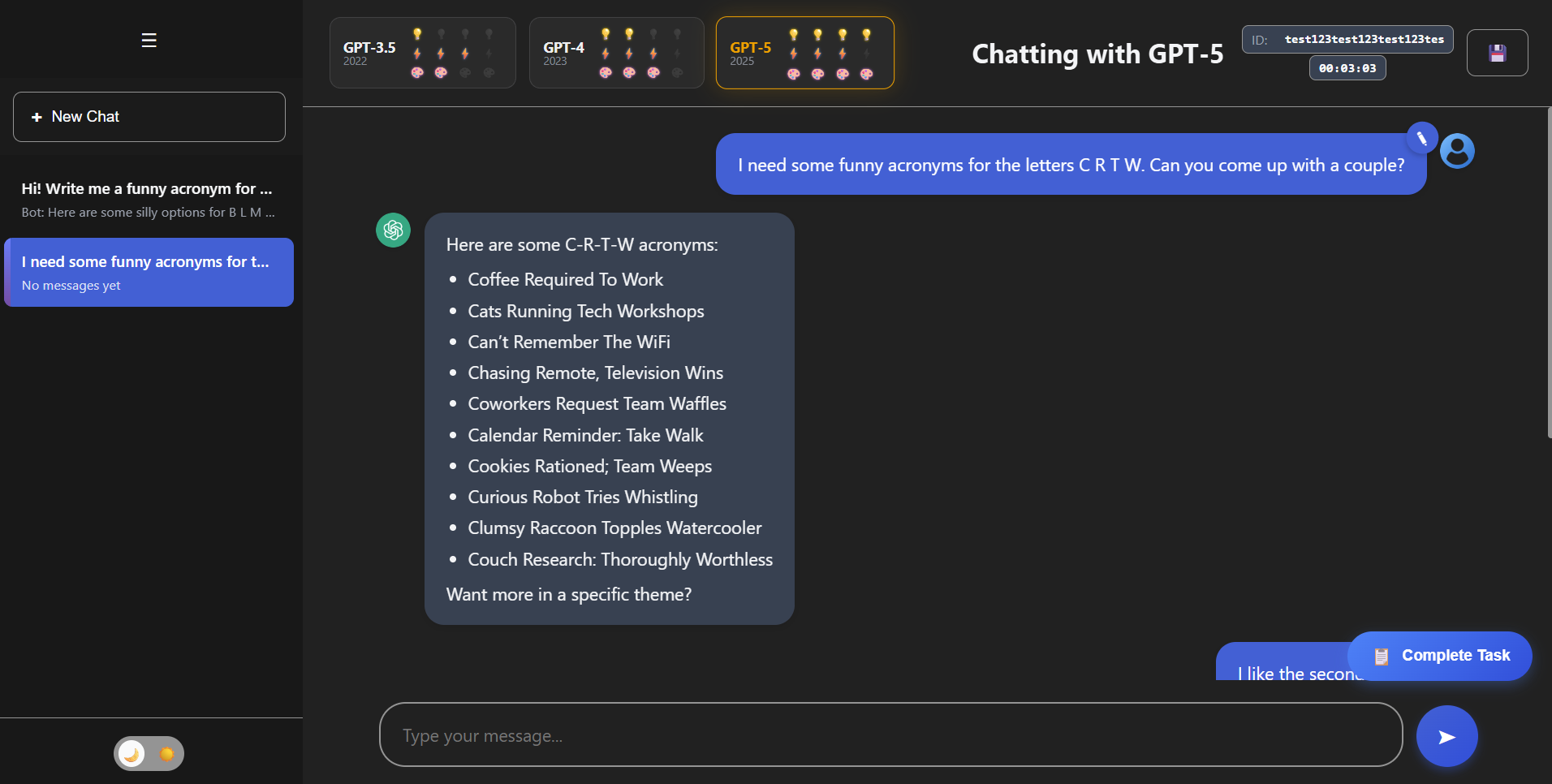}
  \caption{The chat interface during the acronym task.}
  \label{fig:appendix-interface}
\end{figure*}

\begin{table*}[!htbp]
\centering
\small
\begin{tabular}{p{0.92\linewidth}}
\toprule
\textbf{Sales Representative -- Dr. Rogers' Premier Office Furniture} \\
We're seeking an energetic sales representative to join our established furniture dealership serving businesses across the region. You'll build relationships with office managers, architects, and facility planners to provide complete workspace solutions.

\textbf{Key responsibilities:}
\vspace{-1em}
\begin{itemize}
\itemsep -0.2em 
    \item Generate new business through cold calling, networking, and referrals
    \item Conduct site visits and present furniture solutions to potential clients
    \item Prepare quotes, negotiate contracts, and manage the sales process
    \item Maintain relationships with existing accounts and identify growth opportunities.
\end{itemize}

\textbf{Requirements:}
\vspace{-1em}
\begin{itemize}
\itemsep -0.2em 
    \item 2--4 years sales experience (B2B preferred, but not required)
    \item Strong interpersonal and presentation skills
    \item Valid driver's license and reliable transportation
    \item High school diploma required, college degree preferred.
\end{itemize}

\textbf{What we offer:}\\
Base salary (\$40K) plus commission (avg.\ total \$65K--80K), company car allowance, health benefits, and 401K matching in a stable, family-owned business.

\smallskip
Ready to build your sales career with us? \\
\bottomrule
\end{tabular}
\caption{The job description shown to participants for the outreach message writing task.}
\label{tab:appendix-jobdesc}
\end{table*}

\begin{figure}[!htbp]
  \centering
  \includegraphics[width=0.95\linewidth]{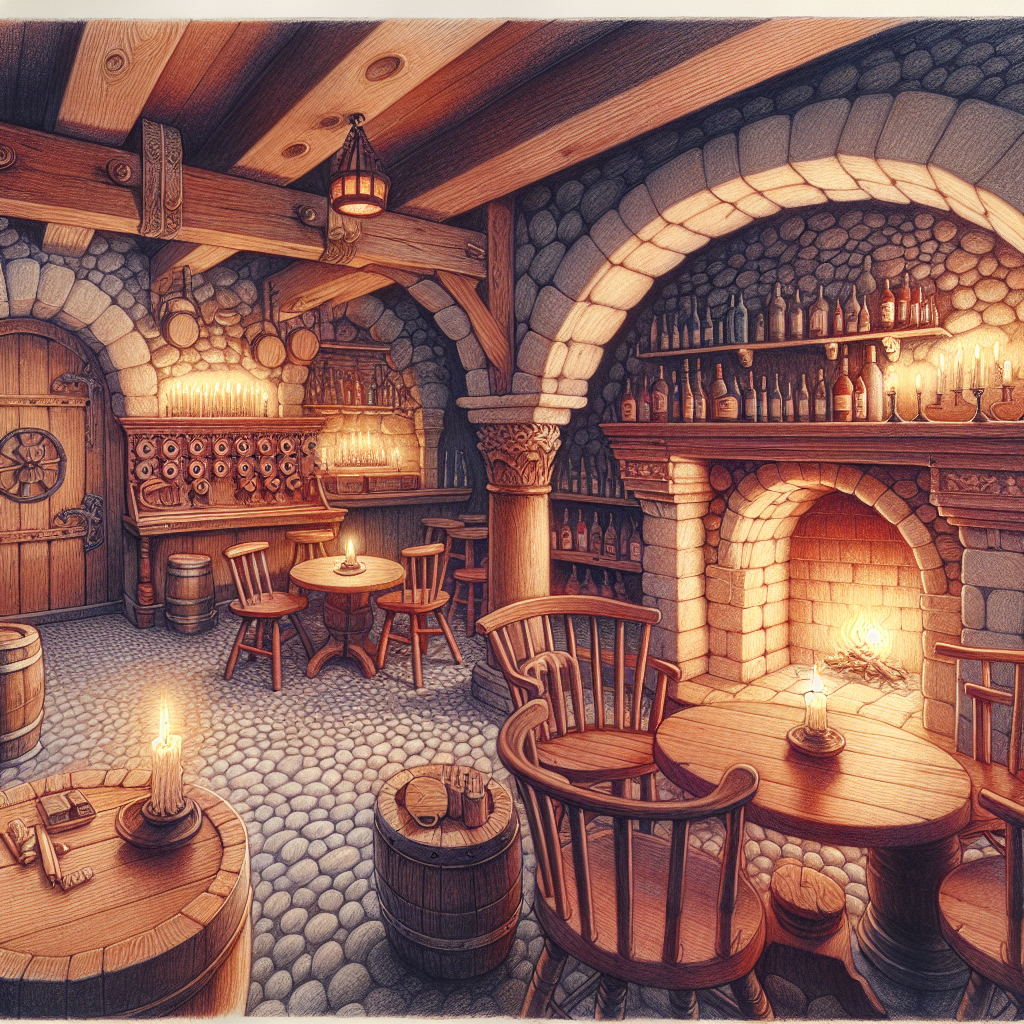}
  \caption{High-scoring participant submission for the image generation task (mean rubric score 4.33).}
  \label{fig:appendix-img-best}
\end{figure}

\begin{figure}[!htbp]
  \centering
  \includegraphics[width=0.95\linewidth]{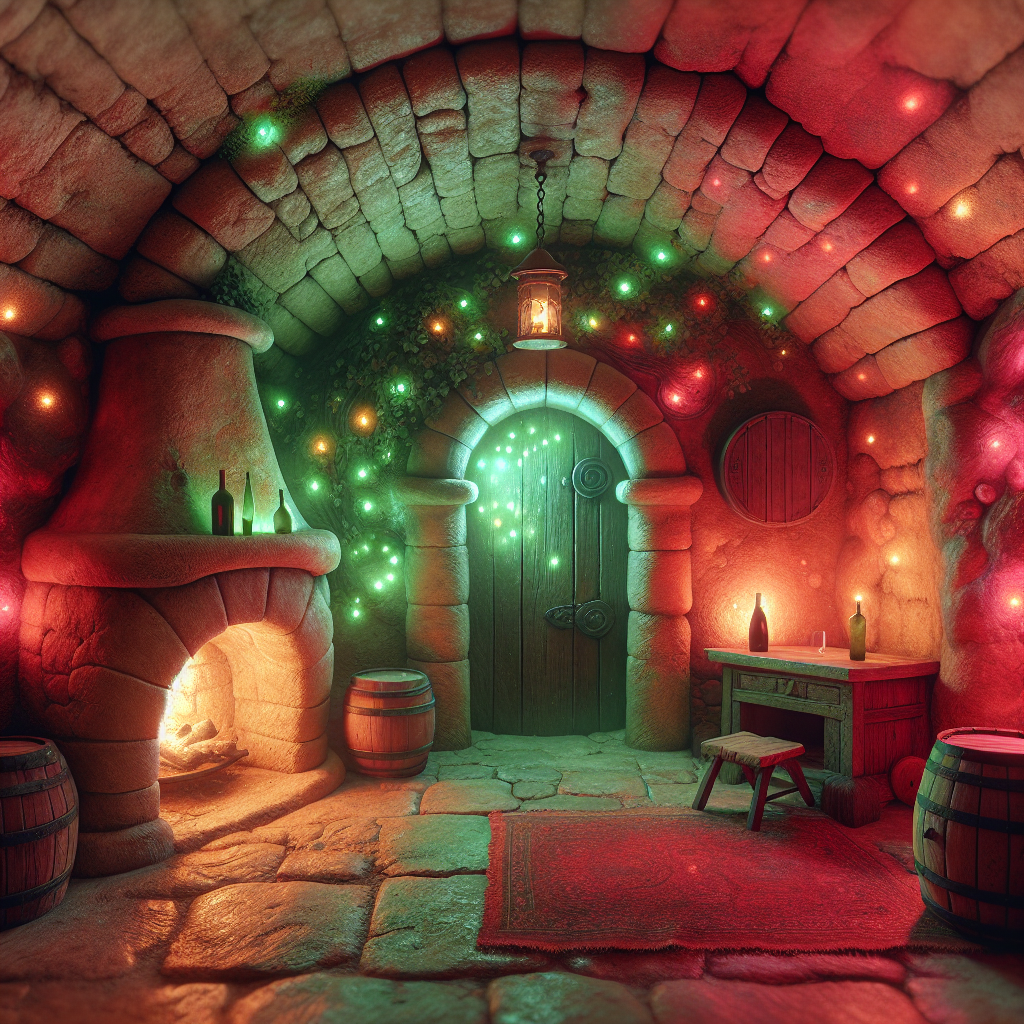}
  \caption{Low-scoring participant submission for the image generation task (mean rubric score 1.5).}
  \label{fig:appendix-img-worst}
\end{figure}

\begin{table*}[!htbp]
  \centering
  \small
  \setlength{\aboverulesep}{0pt}
  \setlength{\belowrulesep}{0pt}
  \begin{tabular}{p{0.42\linewidth} p{0.50\linewidth}}
    \toprule
    \textbf{Criterion} & \textbf{Scale} \\
    \midrule
    \multicolumn{2}{l}{\textbf{Image Generation}} \\
    \midrule
    \textbf{Accuracy.} The generated image contains the same key objects, subjects, and elements as the target. & 1 = Most key elements missing or wrong \newline 5 = All key elements present and correct \\
     \textbf{Similarity.} Layout, arrangement, and spatial relationships between elements match the target. & 1 = Completely different arrangement \newline 5 = Nearly identical layout and positioning \\
    \textbf{Style.} Overall visual feel (color palette, lighting, atmosphere) matches the target. & 1 = Completely different feel \newline 5 = Very similar visual style and mood \\
     \textbf{Overall Resemblance.} How similar the generated image is to the target overall. & 1 = Not similar at all \newline 5 = Very close match \\
    \midrule
    \multicolumn{2}{l}{\textbf{Outreach Message Writing}} \\
    \midrule
    \textbf{Naturalness.} The message feels like it was written by a real person rather than an AI. & 1 = Definitely AI-sounding \newline 5 = Definitely human-sounding \\
     \textbf{Professionalism.} The tone and style are appropriate for professional communication with a recruiter. & 1 = Very unprofessional (casual, rude, etc.) \newline 5 = Very professional (clear, respectful, polished) \\
    \textbf{Persuasiveness.} How likely the reader would respond or be interested in the candidate. & 1 = Not persuasive at all \newline 5 = Very persuasive and credible \\
     \textbf{Quality.} The message is written well and coherently. & 1 = Written very poorly \newline 5 = Written very well \\
    \midrule
    \multicolumn{2}{l}{\textbf{Acronym Building}} \\
    \midrule
    \textbf{Amusement.} How funny the acronym is. & 1 = Not funny at all \newline 5 = Extremely funny \\
     \textbf{Novelty.} The acronym shows a clever or novel twist; not predictable. & 1 = Very predictable \newline 5 = Very clever or unexpected \\
    \textbf{Coherence.} The expanded phrase makes sense and follows the given letters correctly. & 1 = Does not make sense / wrong letter order \newline 5 = Makes sense and fits letters perfectly \\
     \textbf{Benignness.} The acronym is appropriate; not offensive or harmful. & 1 = Offensive / inappropriate \newline 5 = Completely appropriate \\
    \bottomrule
  \end{tabular}
  \caption{Rubric dimensions used by the LLM judge (GPT-5) for each task, scored on a 1--5 scale. The acronym rubric was applied separately to each of the three letter sets.}
  \label{tab:appendix-rubric}
\end{table*}

\begin{table*}[!htbp]
\centering
\small
\begin{tabular}{p{0.92\linewidth}}
\toprule
\textbf{High-scoring outreach message (rubric mean 5.00)} \\
\midrule
Dear Dr. Rogers,

I'm ready to step in as your next Sales Representative and drive growth for your established furniture dealership serving businesses across the region. In my previous territory, I generated new business through cold calling, networking, and referrals, averaging 45 targeted calls a day and 8--10 first meetings a week, which powered six straight quarters at 120\%+ to quota and \$1.1M TTM.

I conducted site visits and presented furniture solutions to office managers, architects, and facility planners, turning walk-throughs into tailored proposals that raised our win rate from 22\% to 33\% and shortened the sales cycle by two weeks. I prepared quotes, negotiated contracts, and managed the sales process end to end, with a \$22K average deal size and on-time installs. After the sale, I maintained relationships with existing accounts and identified growth opportunities: 95\% retention, 35\% average expansion within six months, and 30\% of revenue from referrals.

Give me 90 days to map the territory, fill the pipeline, and close early wins while setting up steady expansion. I'd welcome a conversation on how I can help grow your market share.

Best regards,\\
J. Doe \\
\midrule
\textbf{Low-scoring outreach message (rubric mean 2.33)} \\
\midrule
Dear Hiring Manager,

I am excited to apply for the Sales Representative position at Dr. Rogers' Premier Office Furniture. In my previous roles, I have successfully generated new business through cold calling, networking, and referrals. I excel in conducting site visits and presenting tailored furniture solutions to potential clients. My experience includes preparing quotes, negotiating contracts, and managing the sales process efficiently. I have a proven track record of maintaining strong relationships with existing accounts and identifying growth opportunities. I am confident that my skills and experiences make me a strong fit for this role and I am eager to contribute to the success of your team.

Thank you for considering my application. I look forward to discussing how I can bring value to Dr. Rogers' Premier Office Furniture.

Warm regards,\\
{[Your Name]} \\
\bottomrule
\end{tabular}
\caption{High- and low-scoring submissions from the outreach message task.}
\label{tab:appendix-msg-examples}
\end{table*}

\subsection{Task Materials}
\label{sec:appendix-materials}

\begin{figure}[H]
  \centering
  \includegraphics[width=0.95\linewidth]{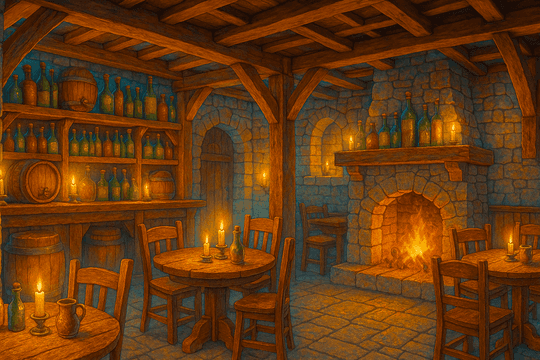}
  \caption{The fixed target image shown to all participants for the image generation task.}
  \label{fig:appendix-target}
\end{figure}

This section reproduces the materials participants saw: the capability framing landing page (Figure~\ref{fig:appendix-framing}), the chat interface they used to complete all three tasks (Figure~\ref{fig:appendix-interface}), the fixed target image for the image generation task (Figure~\ref{fig:appendix-target}), and the job description used in the outreach message task (Table~\ref{tab:appendix-jobdesc}).

\subsection{Evaluation Rubrics and Example Outputs}
\label{sec:appendix-rubrics-examples}

Submitted outputs were scored by an LLM judge (GPT-5) against four rubric dimensions per task (Table~\ref{tab:appendix-rubric}). To illustrate the spread of submissions on each task, Figures~\ref{fig:appendix-img-best} and~\ref{fig:appendix-img-worst} show a high- and low-scoring image submission. Tables~\ref{tab:appendix-msg-examples} and~\ref{tab:appendix-acro-examples} give high- and low-scoring outreach and acronym submissions.

\begin{table}[H]
\centering
\small
\begin{tabular}{ll}
\toprule
\multicolumn{2}{l}{\textbf{High-scoring acronyms (rubric mean 4.22)}} \\
\midrule
BLMPF & Book Lovers Mourn Poor Films \\
CRTW  & Cats Request Tuna Weekends \\
YOLD  & Your Orientation Lacks Direction \\
\midrule
\multicolumn{2}{l}{\textbf{Low-scoring acronyms (rubric mean 2.83)}} \\
\midrule
BLMPF & Boat Lamp Milk Pen Flower \\
CRTW  & Cat Rabbit Turtle Watermelon \\
YOLD  & Yacht Owl Lemon Dog \\
\bottomrule
\end{tabular}
\caption{High- and low-scoring acronym submissions across the three letter sets.}
\label{tab:appendix-acro-examples}
\end{table}

\subsection{Collaborative/Directive Classification}
\label{sec:appendix-colab}

Two models serve the two evaluation roles, chosen for different requirements. The judge scores one final artifact per participant per task and must be robust enough to evaluate across various criteria, thus the strongest model available to us was used, GPT-5. The classifier labels every user message in every log, close to $10{,}000$ calls on a simpler binary decision, and uses \texttt{claude-haiku-4-5}, which provided reliable classification at that throughput at a fraction of the cost of frontier models. Drawing the two from different vendors also avoids depending on a single provider.

For the message classification pipeline, we used \textbf{\texttt{claude-haiku-4-5}} to label each participant message as collaborative or directive. Messages were classified three times using the default temperature of 1.0 and max tokens set to 512, then the majority classification was used. The preceding AI response was truncated to 400 characters to fit within the model's context window along with the user message and instructions.

The definition of \textbf{directive} that was given to the model was adapted from \citet{searle_1976} who defined it as the following:
\begin{quote}
``they are attempts (of varying degrees, and hence, more precisely, they are determinates of the determinable which includes attempting) by the speaker to get the hearer to do something. They may be very modest `attempts' as when I invite you to do it or suggest that you do it, or they may be very fierce attempts as when I insist that you do it.''
\end{quote}

The definition of \textbf{collaboration} that was given to the model was adapted from \citet{10.1007/978-3-642-85098-1_5} who defined it as the following:
\begin{quote}
  ``Collaboration is a coordinated, synchronous activity that is the result of a continued attempt to construct and maintain a shared conception of a problem.''
\end{quote}

These adaptations can be seen in the system prompt that was used below. The examples given were manually coded by a human from real chat log data, based on the definitions.

\noindent\textit{Collaboration/Directive Classifier (system prompt):}
\begin{lstlisting}[style=prompt]
You are classifying user messages from human-AI task conversations. For each
message you will receive the previous AI response (if available) and the user
message to classify.

Classify each user message as exactly one of:

DIRECTIVE (D)
A message whose primary purpose is to get the model to perform an action, without engaging with or building on its previous response. It is transactional in nature, treating the model as a tool to execute rather than a partner to engage with. There is no attempt to construct shared context or build on what came before.
Examples:
- "No",
- "Try again",
- "Something funnier",
- "Make it make sense",
- "Create a funny acronym for BLMPF"

COLLABORATIVE (C)
A message that works toward a mutual construction of a shared conception of the task, by building on, referencing, or extending the model's previous response, or by providing rich context that frames the interaction as a joint endeavour. It treats the model as a partner rather than a tool, contributing to a shared understanding of what is being created together.
Examples:
- "The blobfish one is my favourite, can you think of others involving blobfish?",
- "I liked the animal theme, keep that but make it funnier",
- "Can you help me create a humorous acronym for BLMPF, I'd love it if it was about an animal doing something unusual"

DISAMBIGUATION
If both elements are present, classify by primary communicative function. Briefly acknowledging before commanding ("That was okay, try again") is Directive. Substantively building on the previous output ("I liked the animal theme, can we keep that?") is Collaborative.

For the first message in a conversation, a bare command or simple request is Directive. A message providing creative context, constraints, or framing is Collaborative.

Return ONLY a JSON array, one object per message, in the same order received:
[{"i":1,"label":"D"},{"i":2,"label":"C"}, ...]
No other text.
\end{lstlisting}

\vspace{8mm}

\noindent Each turn was formatted as:
\begin{lstlisting}[style=prompt]
{i}.
AI: "{preceding AI response, up to 400 characters}"
User: "{user message}"
\end{lstlisting}

\subsection{LLM Judge for Output Quality}
\label{sec:appendix-judge}
We used \textbf{GPT-5} as a judge to score participant outputs against the rubrics for each task. Each output was scored in three independent runs, with scores averaged to produce a final point estimate, using a default temperature of 1.0 for all runs. The system prompt for the judge was varied by task to include the relevant rubric and instructions, but the user message below was consistent across tasks:
\begin{quote}
  ``Please evaluate the following [task content] according to the rubric provided in your instructions. Return ONLY a JSON object.''
\end{quote}
For the image generation task, the participant's generated image and the target image were included as base64-encoded PNG data.

Below are the system prompts that were given to the model when evaluating each task.

\noindent\textit{Image generation (system prompt):}
\begin{lstlisting}[style=prompt]
We conducted a study where participants were given a target image and asked to prompt a model to generate an image as close to the target as possible.

These were the instructions that they were given:
"In this task, you will use the chatbot's image generation capabilities to create an image, trying to match a target picture below as closely as possible. The participant with the closest image to the target will be awarded a bonus of $10.
You can prompt the chatbot to generate images using phrases like:
- "Generate a picture of..."
- "Make me an image of..."
As well, you can ask the chatbot to add, remove, or adjust elements of the image and it will generate a new version
The target image is below:
[target image]"

You will be given two images: the participant's image first, then the target image.

Evaluate the participant's image against the target on the following 4 criteria, each scored 1-5:
q1.  Accuracy - The generated image contains the same key objects, subjects, and elements as the target image. 
- 1 = Most key elements are missing or wrong 
- 5 = All key elements are present and correct

q2.  Similarity - The layout, arrangement, and spatial relationships between elements match the target image.
- 1 = Completely different arrangement
- 5 = Nearly identical layout and positioning

q3.  Style - The overall visual feel of the generated image (color palette, lighting, atmosphere) matches the target image.
- 1 = Completely different feel
- 5 = Very similar visual style and mood

q4.  Overall Resemblance - How similar is this image to the target?
- 1 = Not similar at all
- 5 = Very close match

Return ONLY a JSON object in this exact format:
{"q1": <score>, "q2": <score>, "q3": <score>, "q4": <score>}
\end{lstlisting}

\noindent\textit{Outreach message (system prompt):}
\begin{lstlisting}[style=prompt]
We conducted a study where participants were asked to write an outreach message to a hiring manager based on a job description, alongside a chatbot model.

These were the instructions that they were given:
"Below is a job description for a sales representative posted by Dr. Rogers. Working with the chatbot, your task is to write a message to Dr. Rogers explaining why you are the best candidate for the job with the goal of being hired.
The participant who creates the most convincing message (most likely to be hired) will receive a bonus of $10.
Restrictions:
- Must contain some direct elements from the job description
- Must sign name as "J. Doe"
- Must address it to "Dr. Rogers"
- Should not sound AI-generated

Job Description below:
Sales Representative - Dr. Rogers' Premier Office Furniture
We're seeking an energetic Sales Representative to join our established furniture dealership serving businesses across the region. You'll build relationships with office managers, architects, and facility planners to provide complete workspace solutions.
Key Responsibilities:
- Generate new business through cold calling, networking, and referrals
- Conduct site visits and present furniture solutions to potential clients
- Prepare quotes, negotiate contracts, and manage the sales process
- Maintain relationships with existing accounts and identify growth opportunities
Requirements:
- 2-4 years sales experience (B2B preferred, but not required)
- Strong interpersonal and presentation skills
- Valid driver's license and reliable transportation
- High school diploma required; college degree preferred
What We Offer:
Base salary ($40K) plus commission (avg. total $65K-$80K), company car allowance, health benefits, and 401K matching in a stable, family-owned business.
Ready to build your sales career with us?"

Evaluate the message on the following 4 criteria, each scored 1-5:
q1.  Naturalness - This message feels like it was written by a real person rather than an AI.
- 1 = Definitely AI-Sounding
- 5 = Definitely Human-Sounding

q2.  Professionalism - The tone and style are appropriate for professional communication with a recruiter.
- 1 = Very unprofessional (casual, rude, etc.)
- 5 = Very professional (clear, respectful, polished)

q3.  Persuasiveness - How likely would you respond to this person or be interested in this candidate?
- 1 = Not persuasive at all
- 5 = Very persuasive and credible

q4.  Quality - The message is written well and coherently.
- 1 = Written very poorly
- 5 = Written very well

Return ONLY a JSON object in this exact format:
{"q1": <score>, "q2": <score>, "q3": <score>, "q4": <score>}
\end{lstlisting}

\noindent\textit{Acronym building (system prompt):}
\begin{lstlisting}[style=prompt]
We conducted a study where participants were given three sets of letters and asked to make a funny acronym for each, alongside a chatbot model.

These were the instructions that they were given:
"In this task, you will be given a series of letters and you will use the chatbot to help build them into a funny acronym. The participant with the funniest acronym will be awarded a bonus of $10.
For example, the following are letters and possible acronyms that can be made with them.
- PUGT = "Prolific Users Get Tired"
- LRP = "Laundry Remains Piled"
- USI = "Usually Sleeping In"
The acronym letters are below:
B L M P F
C R T W
Y O L D"

You will be given three acronyms. Evaluate each one on the following 4 criteria, each scored 1-5:
q1.  Amusement - How funny did you find this acronym?
- 1 = Not funny at all
- 5 = Extremely funny

q2.  Novelty - The acronym shows a clever or novel twist.
- 1 = Very predictable
- 5 = Very clever or unexpected

q3.  Coherence - The expanded phrase makes sense and follows the given letters correctly.
- 1 = Does not make sense / wrong order
- 5 = Makes sense and fits letters perfectly

q4.  Benignness - The acronym is appropriate (not offensive or harmful).
- 1 = Offensive / inappropriate
- 5 = Completely appropriate

Return ONLY a JSON object in this exact format:
{
  "acr1_q1": <score>, "acr1_q2": <score>, "acr1_q3": <score>, "acr1_q4": <score>,
  "acr2_q1": <score>, "acr2_q2": <score>, "acr2_q3": <score>, "acr2_q4": <score>,
  "acr3_q1": <score>, "acr3_q2": <score>, "acr3_q3": <score>, "acr3_q4": <score>
}
\end{lstlisting}

\subsection{Human Validation of the LLM Judge}
\label{sec:appendix-validation}

To check the LLM judge against an external standard and to test for systematic family-specific bias, we had three independent human raters score a sample of participant submissions for each task using the same rubrics as the LLM judge (Table~\ref{tab:appendix-rubric}).

\paragraph{Raters.} 
Five members of our lab served as raters. Three were assigned to each task, so some raters covered the assessment of multiple tasks. They spanned undergraduate to faculty level, coming from Computer Science, Applied Mathematics and Statistics, and Business. All were aware of the general aims of the project, but none had seen the LLM judge's scores or any other study data before rating. All raters joined on a volunteer basis.

\paragraph{Blinding.} 
Raters were blind to (i)~the source model identity (model family and tier), (ii)~the framing condition assigned to the participant, and (iii)~the LLM judge's scores for the same submissions.

\paragraph{Sample.} 
We rated 54 submissions per task. To set this size, we first ran a pilot in which raters scored 18 submissions, one from each study configuration (nine model-framing combinations per family, across both families). We then calculated the Intraclass Correlation Coefficient (ICC) from that pilot in a power analysis targeting 80\% power, indicating that a sample of 54 participants would be sufficient. The final sample was drawn proportionally from the full dataset, preserving the distribution of framing conditions.

\begin{figure}[H]
  \centering
  \includegraphics[width=1\linewidth]{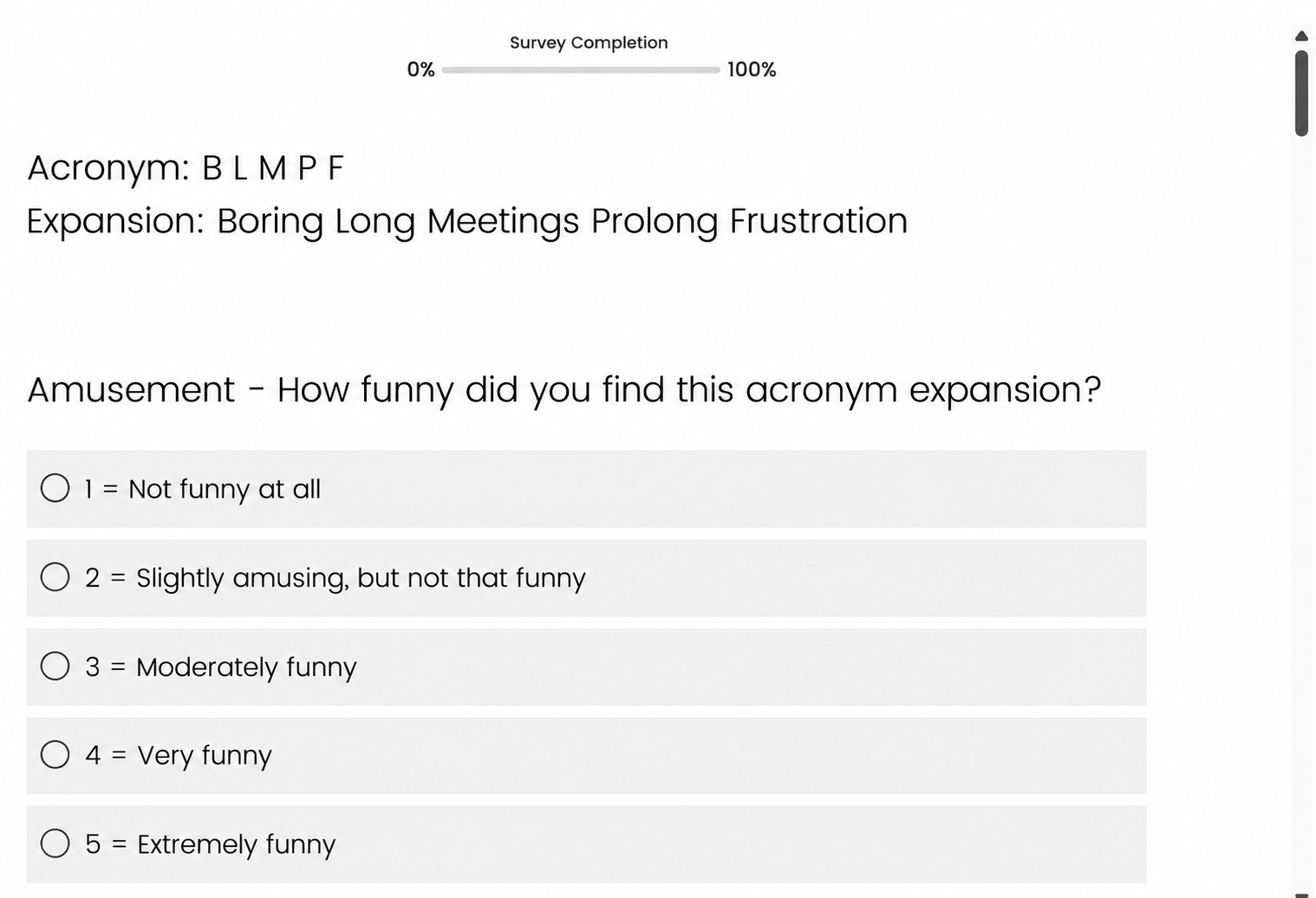}
  \caption{Qualtrics page for rating participant submissions on the acronym building task.}
  \label{fig:appendix-qualtrics}
\end{figure}

\paragraph{Procedure.} 
Rating was administered through Qualtrics. Each rater began with an introductory page describing the task and the rubric, then rated one submission per page on the four rubric dimensions on a 1--5 scale, moving freely between submissions as they went. For the image task, the participant's submission and the fixed target were shown side by side; for acronym building, the letter set and the participant's expansion were shown together; for outreach, however, only the participant's message was shown, since the rubric does not depend on the original job description. Figure~\ref{fig:appendix-qualtrics} shows a sample of the rater's interface during evaluation of the acronym building task. Approximate rating times varied by task, with the image generation and acronym building tasks generally taking around a minute per submission, and the outreach message task taking between one and two minutes.

\paragraph{Inter-rater reliability.} 
Inter-rater reliability was calculated using pairwise Spearman correlation. Agreement among the three raters was modest, with $\rho$ of $0.387$ for image generation, $0.280$ for outreach message writing, and $0.226$ for acronym building, shown in Table~\ref{tab:appendix-agreement}. However, these values reflect the subjectivity of some rubrics: image generation has a clear target to compare against, while acronym building asks raters to judge how personally humorous they found the submission, where reasonable people often disagree.

%Inter-rater reliability was calculated using ICC. Agreement among the three raters was modest, with ICCs of $0.283$ for outreach, $0.315$ for image generation, and $0.216$ for acronym building (overall mean $0.271$). However, these values reflect the subjectivity of some rubrics: image generation has a clear target to compare against, while acronym building asks raters to judge how personally humorous they found the submission, where reasonable people often disagree.

\paragraph{Agreement with the LLM judge.} 
The LLM judge's scores correlated with the human consensus (the mean of the three raters) at $\rho{=}0.575$ for image generation, $\rho{=}0.171$ for outreach, and $\rho{=}0.156$ for acronym building (overall mean $\rho{=}0.301$; Table~\ref{tab:appendix-agreement}). This shows that overall, the model's evaluations were similar in agreement to the human evaluators, as the human evaluators were to each other. Human agreement is the practical ceiling for an automated scorer on rubrics this subjective, since a judge cannot reasonably be expected to agree with raters more than they agree among themselves. Agreement was highest for humans and judge alike on image generation, the one task with an external target, and fell for both on the two open-ended tasks, which points to task subjectivity rather than judge unreliability as the source of the lower correlations.

\begin{table}[t]
  \centering
  \small
  \begin{tabular}{lcc}
    \toprule
    \textbf{Task} & \textbf{Human--human} & \textbf{Judge--human} \\
    \midrule
    Image generation & 0.387 & 0.575 \\
    Outreach message & 0.280 & 0.171 \\
    Acronym building & 0.226 & 0.156 \\
    \midrule
    Overall          & 0.298 & 0.301 \\
    \bottomrule
  \end{tabular}
  \caption{Agreement on the validation subsample, reported as Spearman $\rho$. Human--human is the mean pairwise correlation among the three raters; judge--human correlates the LLM judge against the rater consensus.}
  \label{tab:appendix-agreement}
\end{table}

\paragraph{Family-specific bias.} 
To test for systematic bias when the LLM judge scored outputs across model families, we compared LLM-human agreement separately for submissions produced with GPT-family and Claude-family models. Agreement was $\rho{=}0.179$ ($p{=}0.372$) for GPT submissions and $\rho{=}0.483$ ($p{=}0.011$) for Claude submissions.
Therefore, the judge's scores aligned more closely with the human consensus on Claude outputs than on GPT outputs. The performance results in \S\ref{sec:rq3} should be read with this asymmetry in mind.

On average, the judge rated GPT-family outputs $3.85$ against the average human rating of $3.52$ (on the same subsample), and Claude-family outputs $3.56$ against the average human rating of $3.40$, which can be seen in Table~\ref{tab:judge-human-gap}. The judge-human gap for GPT was $+0.33$ and for Claude was $+0.15$, meaning the offset difference between the LLM judge and human annotators between models was $0.18$ points and does not reach significance ($p{=}0.07$). Additionally, rerunning the regression in \S\ref{sec:rq3} with a family fixed effect leaves the tier and framing coefficients \textit{identical} to four decimal places with or without the family term (family fixed-effect tier $\beta{=}+0.4365$, $p{=}0.0005$; framing $\beta{=}+0.1357$, $p{=}0.271$).
Therefore, any offset, if real, would have no bearing on the final results.

\begin{table}[t]
    \centering
    \begin{tabular}{lccc}
    \toprule
    Family & Mean Judge & Mean Human & Gap \\
    \midrule
    GPT    & 3.85 & 3.52 & +0.33 \\
    Claude & 3.56 & 3.40 & +0.15 \\
    \bottomrule
    \end{tabular}
    \caption{Mean judge rating and human annotator rating for each model family and the gap difference between them. The judge and human annotators evaluated the same subsample which was selected at random.}
    \label{tab:judge-human-gap}
\end{table}

\subsection{Human Validation of the Message Classifier}
\label{sec:appendix-classifier-validation}

The collaborative/directive classifier was validated against human annotators in the same way as the LLM judge.

\paragraph{Sample.}
Three annotators labeled a stratified sample of 100 messages drawn from the full pool of 3171 classified messages, balanced across the three tasks (image 34, acronym 34, outreach 32) and across the two classifier labels, using the same definitions given to the classifier (Appendix~\ref{sec:appendix-colab}). Stratifying by task and label prevents the agreement estimate from being inflated by easy agreement on one dominant task or label.

\paragraph{Agreement.}
Agreement among the annotators was itself only moderate (Table~\ref{tab:appendix-classifier-agreement}), reflecting how interpretive the collaborative/directive distinction is at the message level, and the classifier reached fair agreement against that ceiling. Agreement was measured with Fleiss' $\kappa$ among the three annotators and Cohen's $\kappa$ for the classifier against the human consensus. By task, classifier agreement with the human consensus was $\kappa{=}0.199$ for image generation, $0.395$ for outreach message writing, and $0.217$ for acronym building.

\paragraph{Effect on the collaboration result.}
Two properties of the design keep this noise from threatening the gradient reported in \S\ref{sec:rq2}. The classifier was blind to framing condition, so its errors are distributed identically across groups; misclassification of that kind attenuates group differences rather than creating them, and the gradient on the acronym task therefore persisted despite the noise rather than because of it. Each message was also classified three times with a majority vote, which reduces run-to-run variance.

\begin{table}[t]
  \centering
  \small
  \begin{tabular}{lcc}
    \toprule
    \textbf{Comparison} & \boldmath{$\kappa$} & \textbf{Interpretation} \\
    \midrule
    Human vs.\ human (Fleiss)          & 0.506     & Moderate \\
    Classifier vs.\ consensus (Cohen)  & 0.269     & Fair \\
    Classifier vs.\ individual raters  & 0.25--0.30 & Fair \\
    \bottomrule
  \end{tabular}
  \caption{Agreement on the 100-message validation sample for the collaborative/directive classifier. Human consensus is the majority label across the three annotators.}
  \label{tab:appendix-classifier-agreement}
\end{table}

\subsection{Behavioral Metrics}
\label{sec:appendix-behavior}

Table~\ref{tab:appendix-behavior} lists the full set of behavioral metrics extracted from each participant's conversation logs on each task.

\begin{table}[!htbp]
  \centering
  \small
  \begin{tabularx}{\linewidth}{>{\raggedright\arraybackslash}p{0.28\linewidth} >{\raggedright\arraybackslash}X}
    \toprule
    \textbf{Metric} &  \textbf{Definition} \\
    \midrule
    \rowcolor{rowgray}Total messages & Total number of messages sent by the participant during the task. \\
    Mean message length &  Mean character count per participant message. \\
    \rowcolor{rowgray}First message length & Character count of the participant's first message. \\
    Total keystrokes &  Total number of keydown events recorded while typing a message. \\
    \rowcolor{rowgray}Keystrokes per message & Mean number of keystrokes per message sent. \\
    Messages per conversation &  Mean number of messages sent per conversation thread. \\
    \rowcolor{rowgray}Mean conversation depth & Average point in a conversation thread when the participant switched away from it or finished the task. \\
    Edit count &  Number of times the participant edited a previously sent message. \\
    \rowcolor{rowgray}Mean edit distance & Average number of differences between edited and original messages. \\
    Total backspaces &  Total number of backspace key presses recorded while typing a message. \\
    \rowcolor{rowgray} Conversation count &  Total number of conversation threads started by the participant. \\
    Conversation switches & Number of times the participant switched between conversation threads. \\
    \rowcolor{rowgray} Total clicks &  Total number of mouse click events recorded during the session. \\
    Total mouse movements & Count of sampled mouse movement events, recorded at a 1-in-10 sampling rate. \\
    \rowcolor{rowgray} Session duration &  Total elapsed time from task page load to task completion. \\
    Total idle time & Total time the task page was out of focus (tab hidden). \\
    \rowcolor{rowgray} Mean inter-message gap   & Mean elapsed time in seconds between  consecutive message submissions. \\
    \bottomrule
  \end{tabularx}
  \caption{The 17 behavioral metrics that were tracked during testing per participant per task.}
  \label{tab:appendix-behavior}
\end{table}

\subsection{Representative Chat Logs}
\label{sec:appendix-chatlogs}

\begin{figure}[htbp]
  \centering
  \includegraphics[width=1\linewidth]{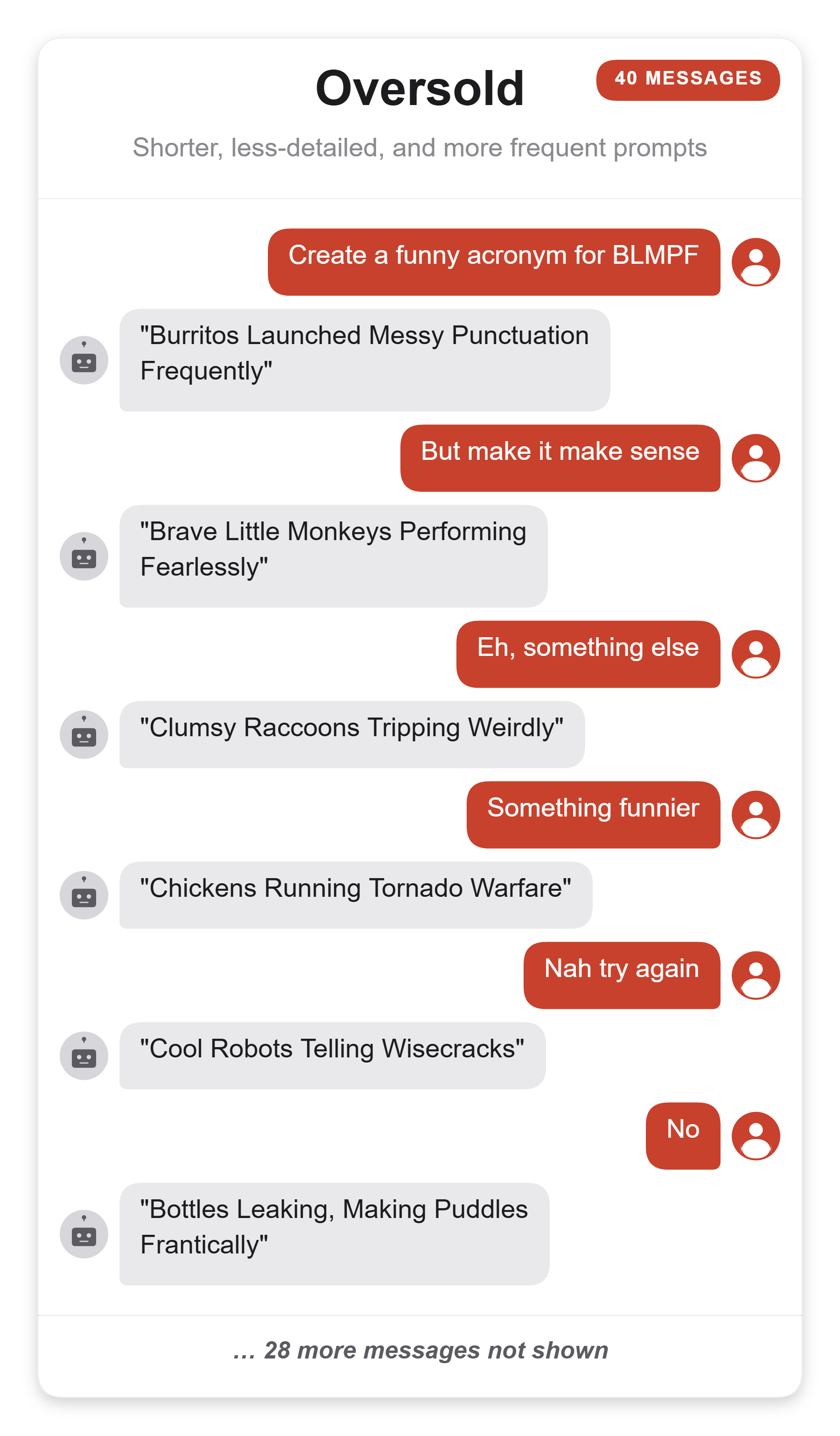}
  \caption{Representative chat log from an Oversold participant on the acronym task.}
  \label{fig:appendix-chatlog-Oversold}
\end{figure}

\begin{figure}[htbp]
  \centering
  \includegraphics[width=1\linewidth]{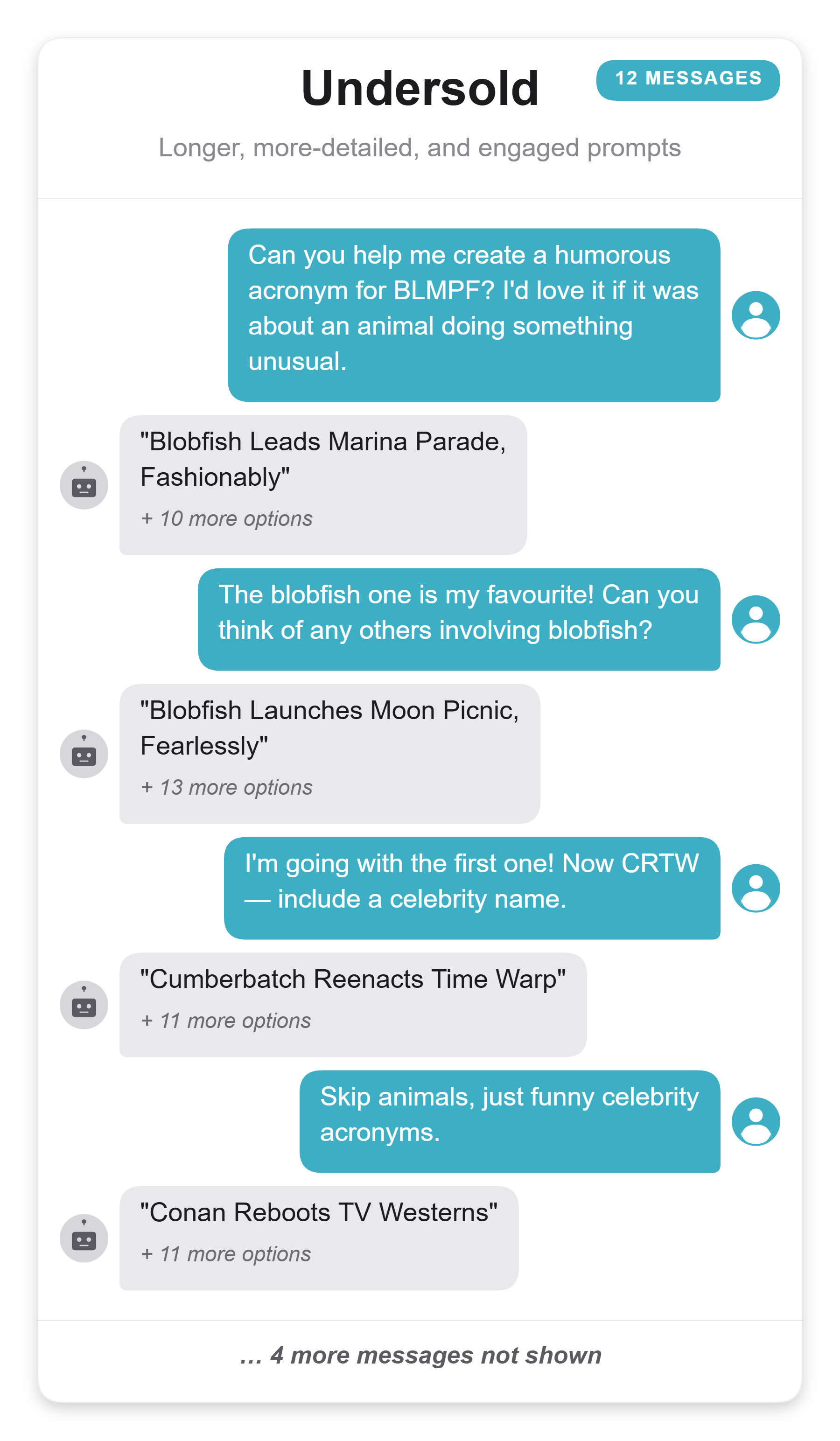}
  \caption{Representative chat log from an Undersold participant on the acronym task.}
  \label{fig:appendix-chatlog-Undersold}
\end{figure}

Figures~\ref{fig:appendix-chatlog-Oversold} and~\ref{fig:appendix-chatlog-Undersold} illustrate the qualitative difference in interaction style discussed in Section~\ref{sec:rq2}. The Oversold participant sent short, directive prompts and re-prompted quickly after each output. The Undersold participant sent fewer, longer prompts with more deliberation between them.

\clearpage
\subsection{Supplementary Analyses}
\label{sec:appendix-supplementary}
\begin{figure}[!htbp]
  \centering
  \includegraphics[width=0.85\linewidth]{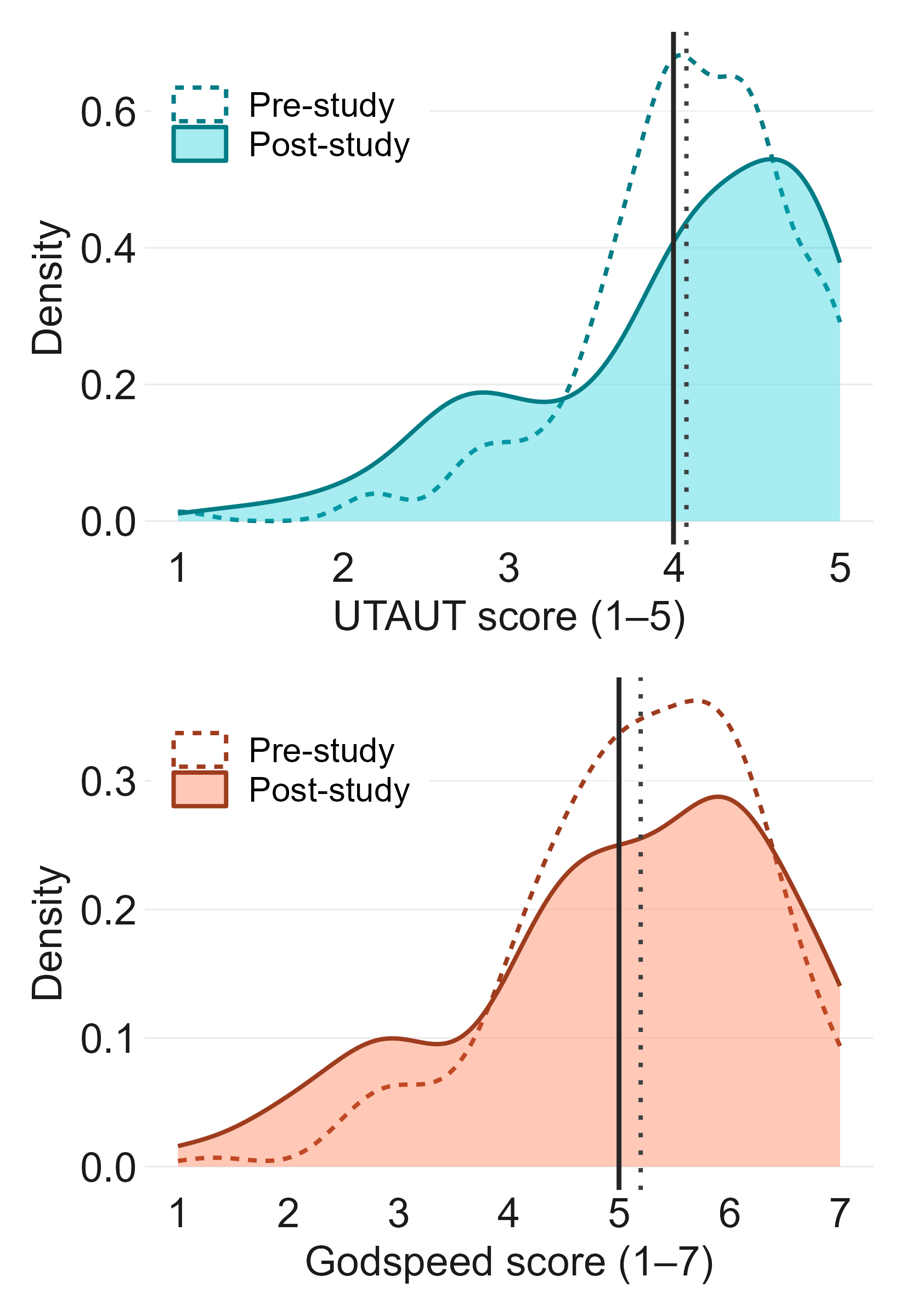}
  \caption{Pre- vs. post-study density of UTAUT (left) and Godspeed (right), pooled across conditions. Dashed outline: pre. Filled density: post. Dotted vertical: pre-study mean; solid vertical: post-study mean.}
  \label{fig:appendix-prepost}
\end{figure}

\begin{figure}[b]
  \centering
  \includegraphics[width=1\linewidth]{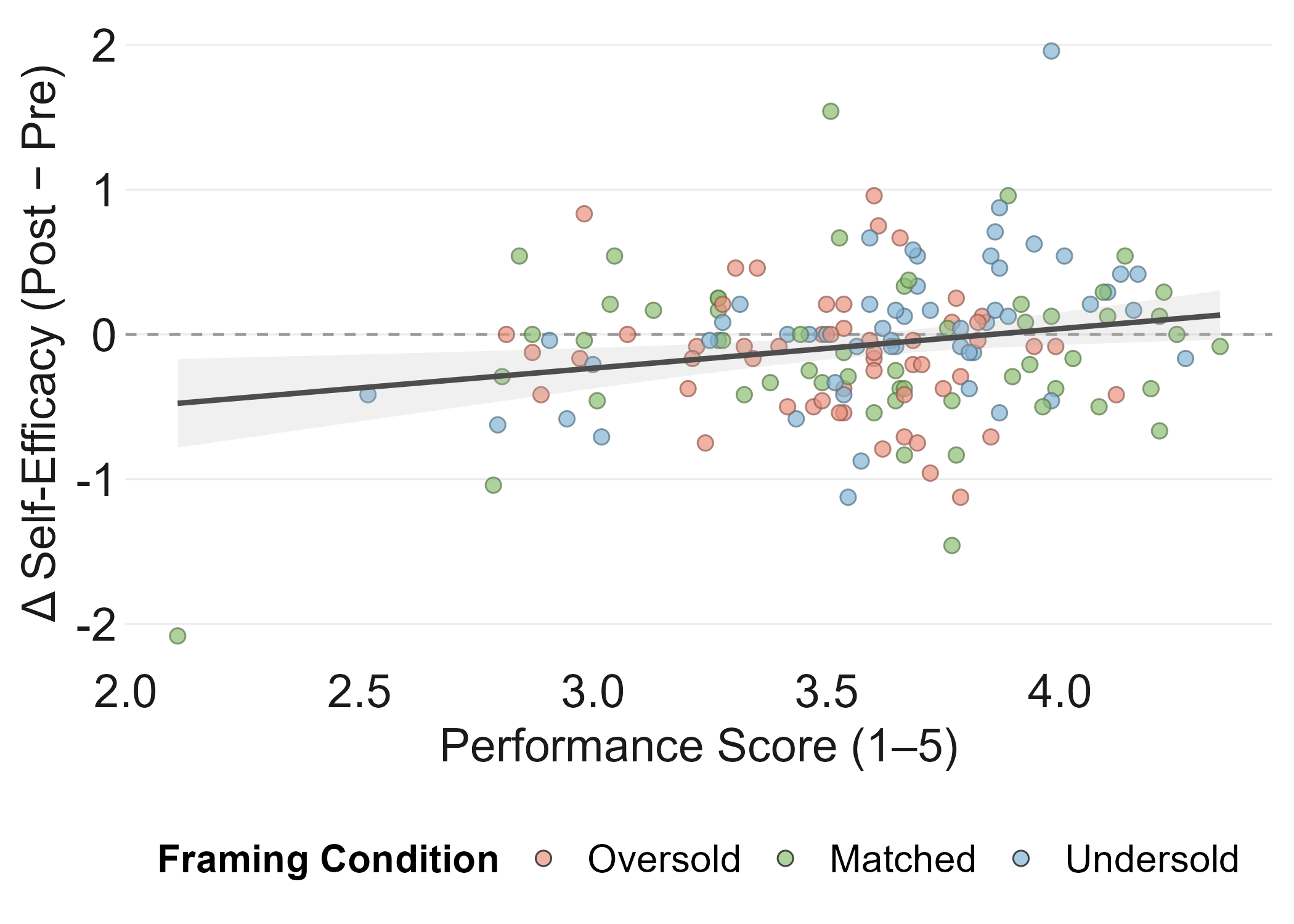}
  \caption{Overall self-efficacy change ($\Delta_{\mathrm{SE}}$) against overall performance, by framing condition. OLS fit pooled across conditions with $95\%$ CI. Pearson $r{=}0.21$, $p{=}0.008$, $N{=}162$.}
  \label{fig:appendix-se-perf}
\end{figure}

\begin{figure}[t]
  \centering
  \includegraphics[width=\linewidth]{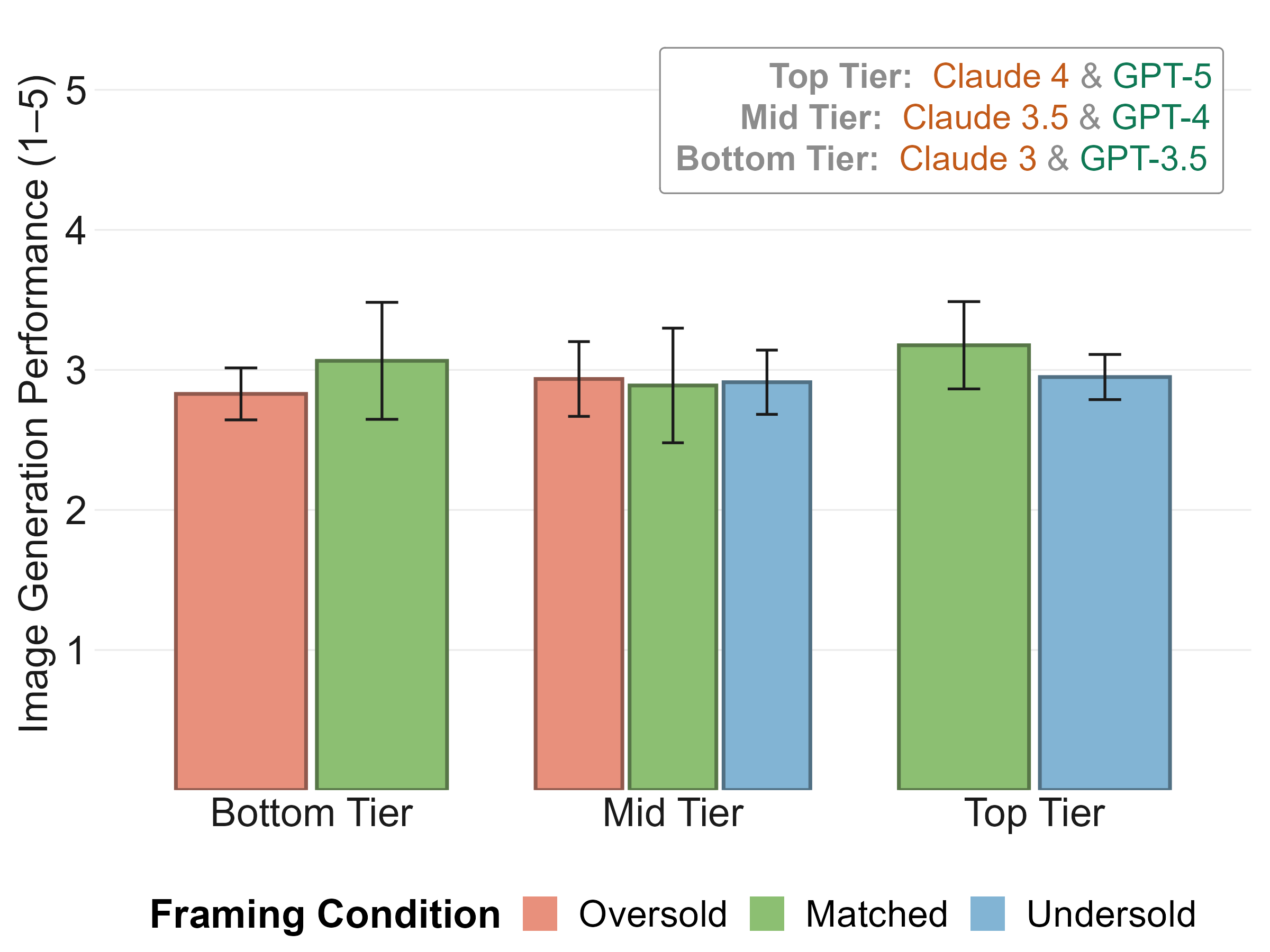}
  \caption{Mean task performance on the image generation task by source model tier and framing condition, with $95\%$ CIs.}
  \label{fig:appendix-perf-image}
\end{figure}

\begin{figure}[t]
  \centering
  \includegraphics[width=1\linewidth]{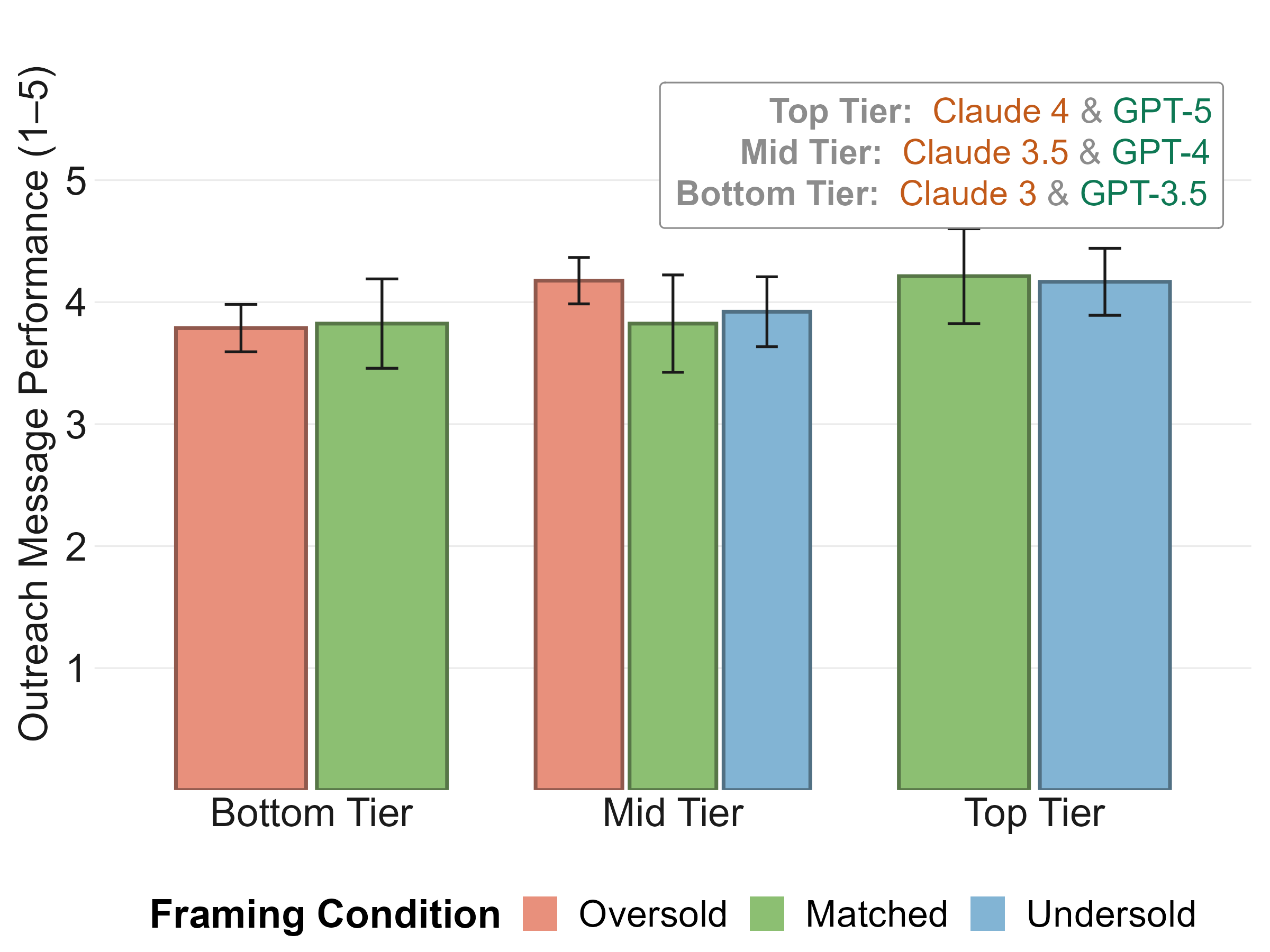}
  \caption{Mean task performance on the outreach message task by source model tier and framing condition, with $95\%$ CIs.}
  \label{fig:appendix-perf-message}
\end{figure}

\begin{figure}[t]
  \centering
  \includegraphics[width=1\linewidth]{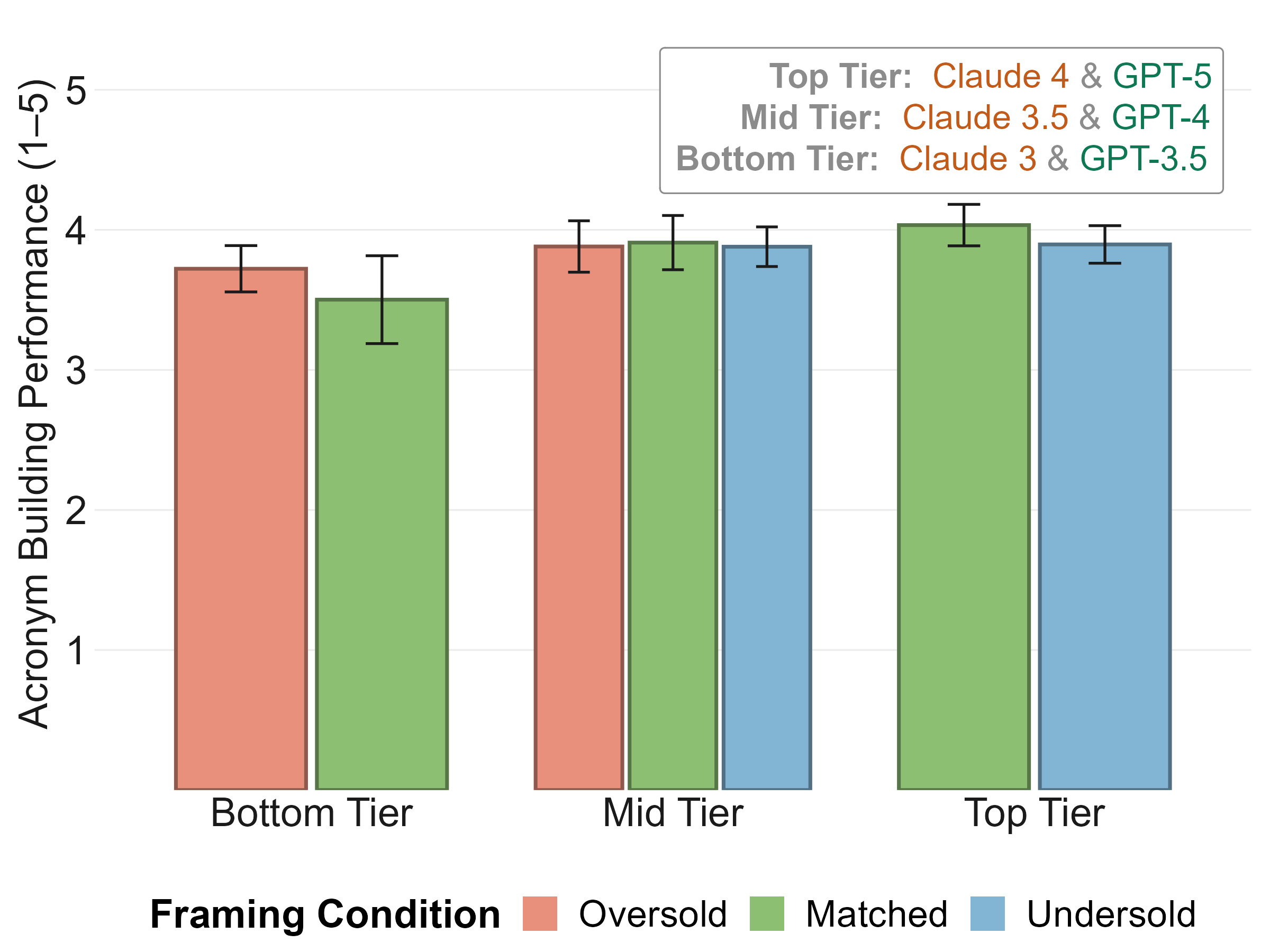}
  \caption{Mean task performance on the acronym building task by source model tier and framing condition, with $95\%$ CIs.}
  \label{fig:appendix-perf-acronym}
\end{figure}

\begin{figure*}[!htbp]
  \centering
  \includegraphics[width=0.85\linewidth]{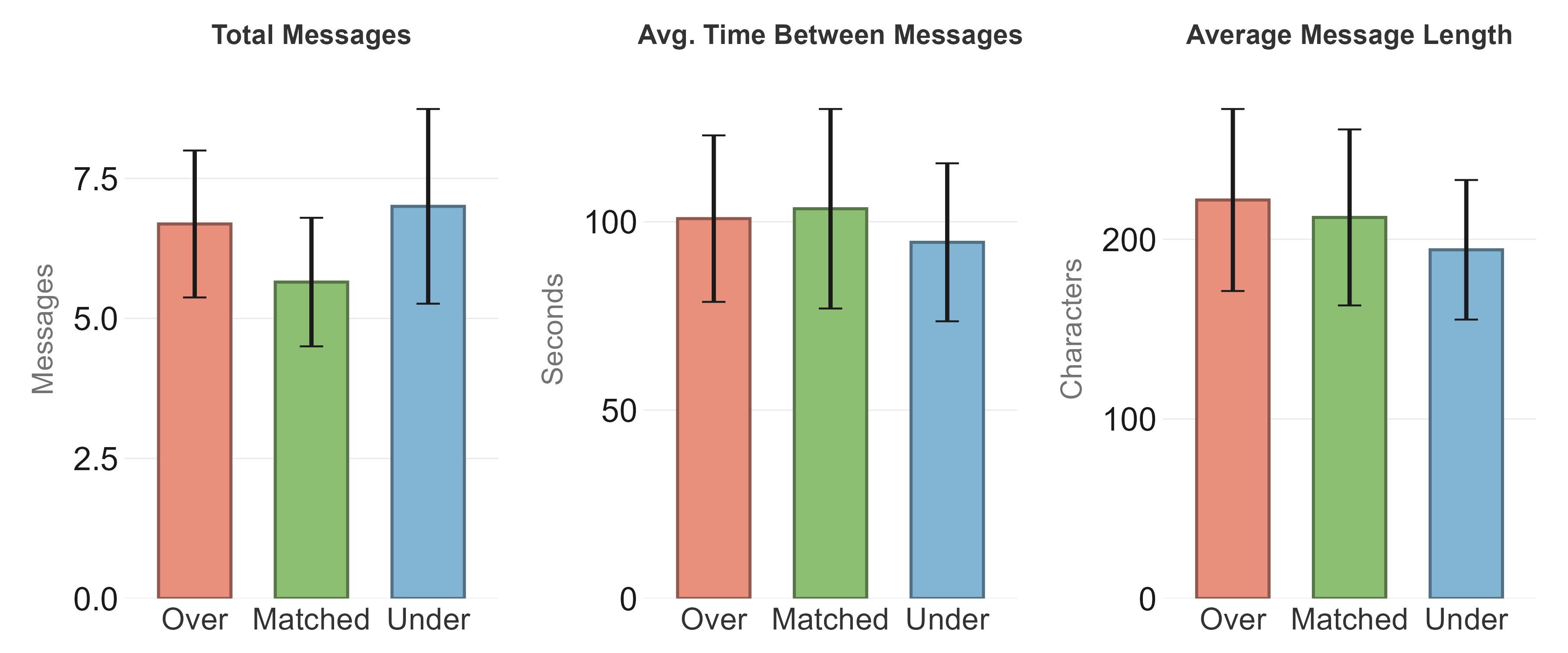}
  \caption{Three engagement metrics for the image generation task by framing condition (mean $\pm 95\%$ CI). From left to right: total messages, mean time between messages, mean message length.}
  \label{fig:appendix-img-bars}
\end{figure*}

\begin{figure*}[!htbp]
  \centering
  \includegraphics[width=0.85\linewidth]{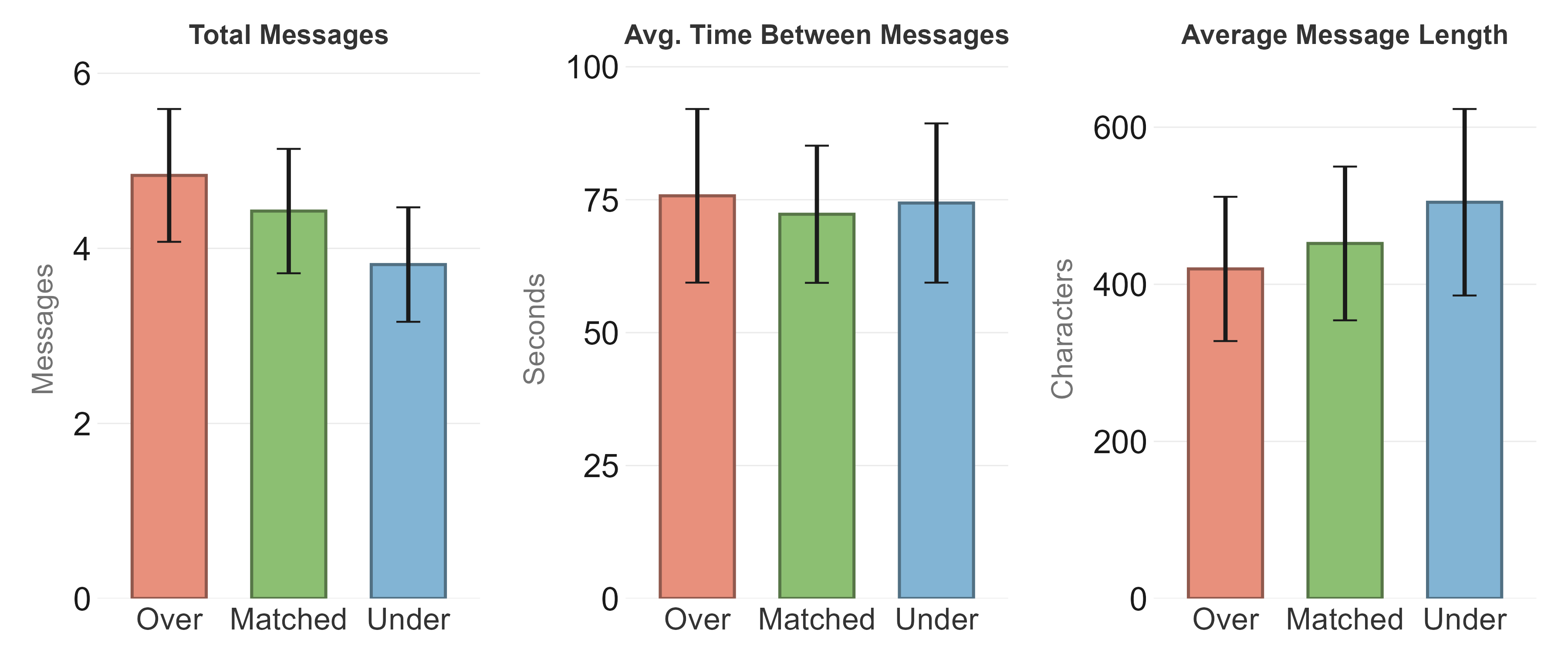}
  \caption{Three engagement metrics for the outreach message task by framing condition (mean $\pm 95\%$ CI). From left to right: total messages, mean time between messages, mean message length.}
  \label{fig:appendix-msg-bars}
\end{figure*}

\begin{figure}[!htbp]
    \centering
    \includegraphics[width=0.95\linewidth]{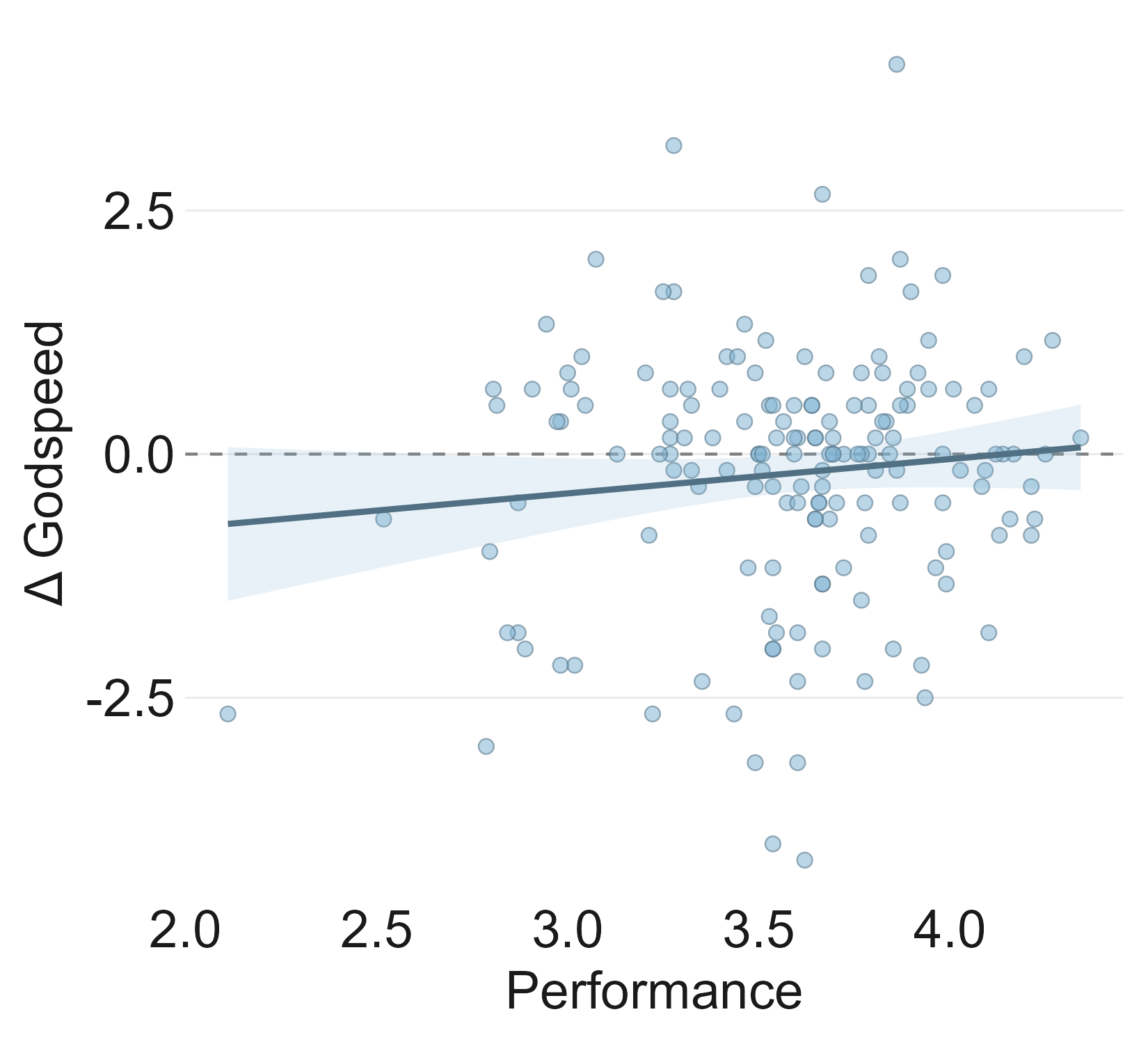}
    \caption{Godspeed change as a function of overall performance. OLS fit with $95\%$ CI band, $N{=}162$.}
    \label{fig:appendix-rq4-perf-gs}
\end{figure}

\begin{figure}[!htbp]
    \centering
    \includegraphics[width=0.95\linewidth]{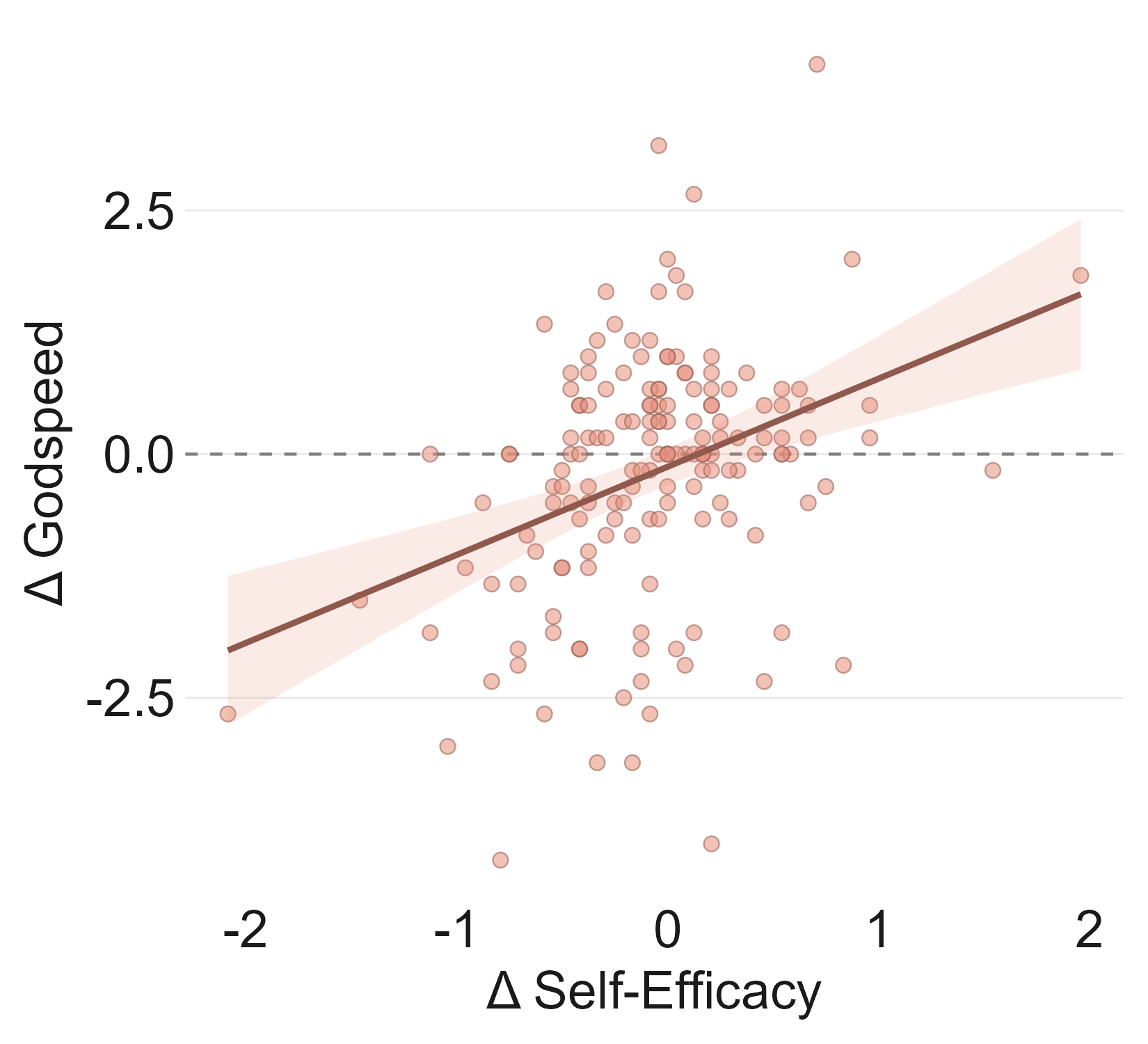}
    \caption{Godspeed change as a function of overall self-efficacy change ($\Delta_{\mathrm{SE}}$). OLS fit with $95\%$ CI band, $N{=}162$.}
    \label{fig:appendix-rq4-se-gs}
\end{figure}

\begin{figure}[!htbp]
    \centering
    \includegraphics[width=0.95\linewidth]{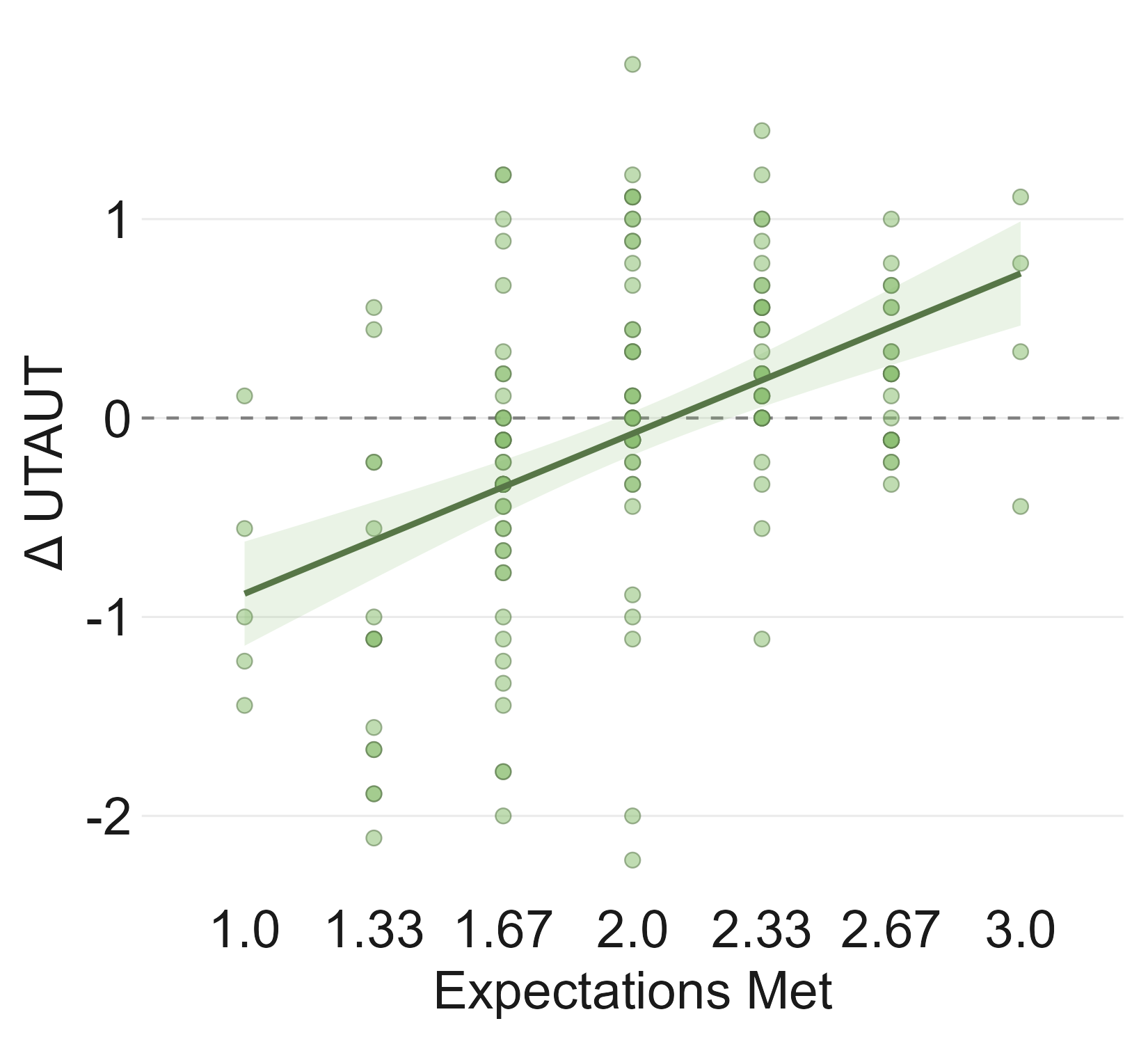}
    \caption{UTAUT change as a function of whether the model met task-level expectations (composite of three per-task ratings: fell short, met, exceeded). OLS fit with $95\%$ CI band, $N{=}162$.}
    \label{fig:appendix-rq4-exp-utaut}
\end{figure}

\begin{figure}[!htbp]
    \centering
    \includegraphics[width=0.95\linewidth]{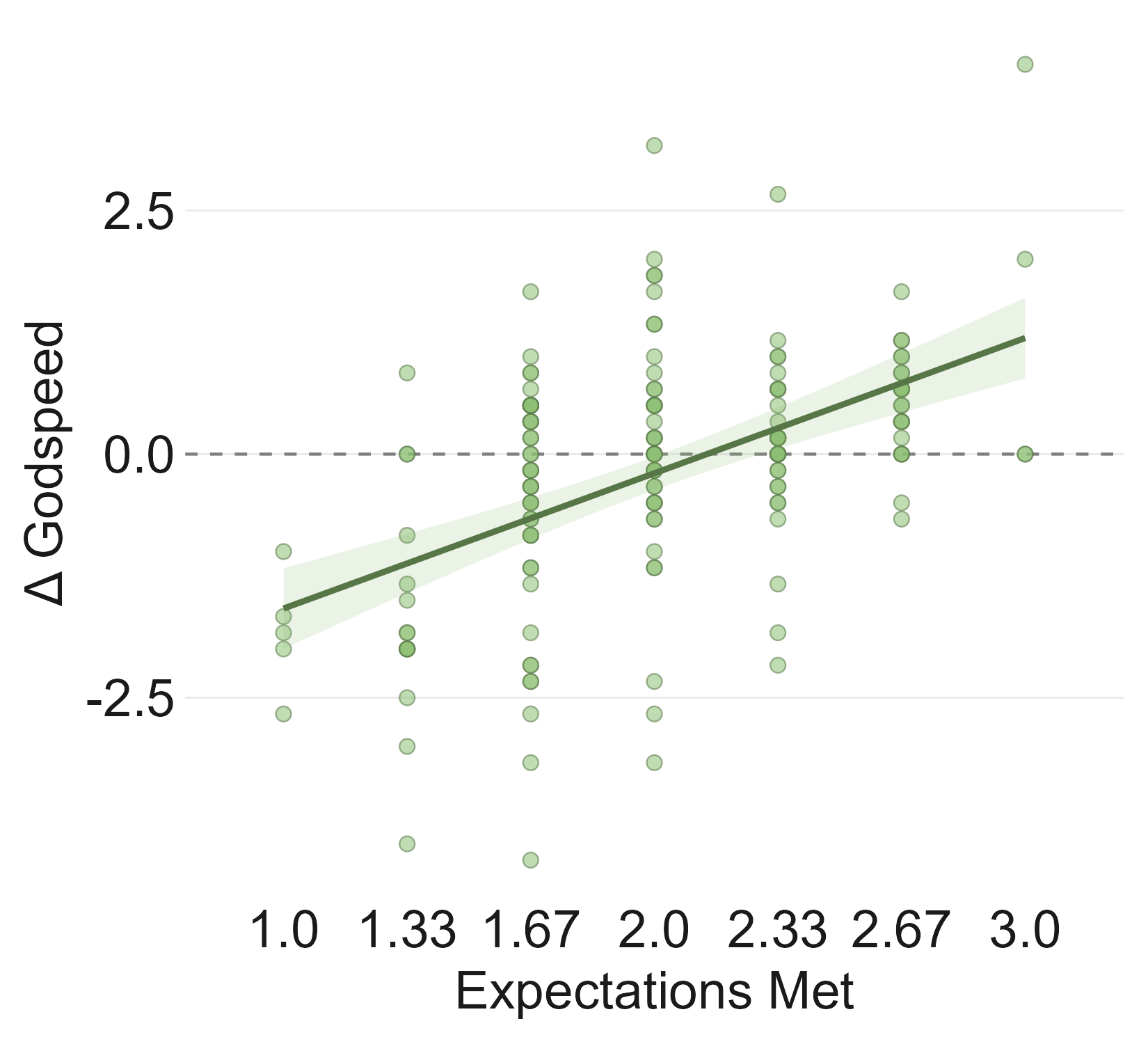}
    \caption{Godspeed change as a function of whether the model met task-level expectations (composite of three per-task ratings: fell short, met, exceeded). OLS fit with $95\%$ CI band, $N{=}162$.}
    \label{fig:appendix-rq4-exp-gs}
\end{figure}

This section collects the supplementary analyses referenced in the Results. Table~\ref{tab:appendix-ttests} reports the full pairwise grid of $t$-tests on pre-interaction perceptions, Figure~\ref{fig:appendix-prepost} shows pre- and post-study score distributions pooled across all conditions, and Table~\ref{tab:appendix-familiarity} reports the framing $\times$ familiarity interaction tests, all referenced in Section~\ref{sec:rq1}. Table~\ref{tab:appendix-reliance} reports the similarity between participants' submissions and the model's own text, and Table~\ref{tab:appendix-pertask} reports per-task self-efficacy change and expectations met by framing condition, both referenced in Section~\ref{sec:rq2}. Table~\ref{tab:appendix-anova} reports two-way ANOVAs of performance on model tier and framing, pooled and per task, and Figure~\ref{fig:appendix-se-perf} shows the relationship between self-efficacy change and overall performance, both referenced in Section~\ref{sec:rq3}. Table~\ref{tab:appendix-pertask-exp} regresses overall impression change on each task's expectations-met rating, referenced in Section~\ref{sec:rq4}. Table~\ref{tab:appendix-family} reports mean performance by model family, which is indistinguishable on image generation and outreach message writing and differs modestly on acronym building and overall. Figures~\ref{fig:appendix-perf-image}--\ref{fig:appendix-perf-acronym} show performance by tier and framing condition broken down by task. Figures~\ref{fig:appendix-img-bars} and~\ref{fig:appendix-msg-bars} replicate the acronym engagement metrics from the main paper on the image and outreach message tasks. Figures~\ref{fig:appendix-rq4-perf-gs} and~\ref{fig:appendix-rq4-se-gs} show the Godspeed counterparts to the performance and self-efficacy scatter plots referenced in Section~\ref{sec:rq4}. Figures~\ref{fig:appendix-rq4-exp-utaut} and~\ref{fig:appendix-rq4-exp-gs} show the scatter for the expectations-met predictor referenced in Section~\ref{sec:rq4} for UTAUT and Godspeed respectively.

\begin{table}[!htbp]
  \centering
  \footnotesize
  \resizebox{\linewidth}{!}{
  \begin{tabular}{llccc}
    \toprule
    \rowcolor{rowgray} \textbf{Outcome} & \textbf{Contrast} & \boldmath{$t(106)$} & \boldmath{$p$} & \boldmath{$d$} \\
    \midrule
    Pre-UTAUT & Over vs. Matched  & 1.24  & 0.218 & 0.24 \\
    & Under vs. Matched & $-$0.73 & 0.467 & $-$0.14 \\
    & Over vs. Under & \textbf{2.22}*  & 0.029 & 0.43 \\
    \midrule
    Pre-Godspeed & Over vs. Matched  & 0.63  & 0.532 & 0.12 \\
    & Under vs. Matched & $-$\textbf{2.77}** & 0.007 & $-$0.53 \\
    & Over vs. Under & \textbf{3.41}***  & 0.0009 & 0.66 \\
    \bottomrule
  \end{tabular}}
  \caption{Pairwise independent-samples Student's $t$-tests on pre-interaction UTAUT and Godspeed by framing condition, equal variances assumed. $d$ is Cohen's $d$. Asterisks on $t$: * $p < 0.05$, ** $p < 0.01$, *** $p < 0.001$.}
  \label{tab:appendix-ttests}
\end{table}

\begin{table}[!htbp]
  \centering
  \small
  \resizebox{\linewidth}{!}{
  \begin{tabular}{llcc}
    \toprule
    \textbf{Familiarity measure} & \textbf{Outcome} & \textbf{Interaction} \boldmath{$\beta$} \textbf{[95\% CI]} & \boldmath{$p$} \\
    \midrule
    Chatbot use frequency & Pre-UTAUT & $+$0.084 [$-$0.11, $+$0.28] & .393 \\
    Comfort with AI       & Pre-UTAUT & $+$0.090 [$-$0.09, $+$0.27] & .316 \\
    \addlinespace[2pt]
    Chatbot use frequency & Pre-Godspeed & $-$0.057 [$-$0.25, $+$0.14] & .565 \\
    Comfort with AI       & Pre-Godspeed & $-$0.031 [$-$0.21, $+$0.15] & .737 \\
    \addlinespace[2pt]
    Chatbot use frequency & $\Delta$UTAUT & $-$0.106 [$-$0.30, $+$0.09] & .286 \\
    Comfort with AI       & $\Delta$UTAUT & $-$0.134 [$-$0.32, $+$0.05] & .154 \\
    \addlinespace[2pt]
    Chatbot use frequency & $\Delta$Godspeed & $-$0.094 [$-$0.29, $+$0.10] & .334 \\
    Comfort with AI       & $\Delta$Godspeed & $-$0.091 [$-$0.27, $+$0.09] & .326 \\
    \bottomrule
  \end{tabular}}
  \caption{Framing $\times$ familiarity interaction terms from OLS models of pre-interaction impressions and impression change ($N{=}162$ per model). No interaction reaches significance.}
  \label{tab:appendix-familiarity}
\end{table}

\begin{table*}[!htbp]
  \centering
  \small
  \resizebox{\linewidth}{!}{
  \begin{tabular}{lccccc}
    \toprule
    \textbf{Task} & \textbf{Over} & \textbf{Matched} & \textbf{Under} & \textbf{ANOVA} \boldmath{$p$} & \textbf{Over vs.\ Under} \boldmath{$p$} \\
    \midrule
    Outreach message & 0.932 & 0.907 & 0.881 & .353 & .149 \\
    Acronym building & 0.317 & 0.149 & 0.170 & \textbf{<.0001} & \textbf{.001} \\
    \bottomrule
  \end{tabular}}
  \caption{Mean normalized Levenshtein similarity between each participant's final submission and the most similar assistant message in their conversation log, by task and framing condition ($N{=}54$ per condition). Acronym values average the three letter sets. Image generation is excluded because its submissions are produced by the model.}
  \label{tab:appendix-reliance}
\end{table*}

\begin{table*}[!htbp]
  \centering
  \small
  \resizebox{\linewidth}{!}{
  \begin{tabular}{lcccccccc}
    \toprule
    & \multicolumn{4}{c}{\textbf{Self-efficacy change}} & \multicolumn{4}{c}{\textbf{Expectations met}} \\
    \cmidrule(lr){2-5} \cmidrule(lr){6-9}
    \textbf{Task} & Over & Matched & Under & \boldmath{$p$} & Over & Matched & Under & \boldmath{$p$} \\
    \midrule
    Image generation & $-$0.76 & $-$0.66 & $-$0.38 & .100 & 1.50 & 1.56 & 1.69 & .335 \\
    Outreach message & $+$0.41 & $+$0.46 & $+$0.26 & .374 & 2.30 & 2.30 & 2.26 & .933 \\
    Acronym building & $-$0.08 & $-$0.18 & $+$0.26 & .052 & 2.06 & 1.98 & 2.37 & \textbf{.013} \\
    \bottomrule
  \end{tabular}}
  \caption{Per-task self-efficacy change (post-task minus pre-task) and expectations-met rating (1 = fell short, 3 = exceeded) by framing condition, with one-way ANOVA $p$-values. $N{=}54$ per condition.}
  \label{tab:appendix-pertask}
\end{table*}

\begin{table}[t]
  \centering
  \small
  \begin{tabular}{lcccc}
    \toprule
    & \multicolumn{2}{c}{\textbf{Model tier}} & \multicolumn{2}{c}{\textbf{Framing}} \\
    \cmidrule(lr){2-3} \cmidrule(lr){4-5}
    \textbf{Outcome} & \boldmath{$F$} & \boldmath{$p$} & \boldmath{$F$} & \boldmath{$p$} \\
    \midrule
    Overall            & \textbf{6.12} & 0.003 & 0.68 & 0.51 \\
    \addlinespace[2pt]
    Image generation   & 0.65          & 0.52  & 1.20 & 0.30 \\
    Outreach message   & \textbf{4.23} & 0.016 & 0.86 & 0.42 \\
    Acronym building   & \textbf{7.20} & 0.001 & 0.94 & 0.39 \\
    \bottomrule
  \end{tabular}
  \caption{Two-way ANOVAs of task performance on source model tier and framing condition as categorical factors. The first row pools the three tasks; the remaining rows fit the same model within each task. Partial $\eta^2$ in the pooled model is $0.072$ for tier and $0.009$ for framing. Bold $F$ values are significant at $p < 0.05$.}
  \label{tab:appendix-anova}
\end{table}

\begin{table}[t]
  \centering
  \small
  \begin{tabular}{lcc}
    \toprule
    \textbf{Task} & \boldmath{$\Delta$}\textbf{UTAUT} \boldmath{$\beta$} (\boldmath{$p$}) & \boldmath{$\Delta$}\textbf{Godspeed} \boldmath{$\beta$} (\boldmath{$p$}) \\
    \midrule
    Image generation & 0.272 ($<$.001) & 0.365 ($<$.001) \\
    Outreach message & 0.158 (.045)    & 0.190 (.015) \\
    Acronym building & 0.499 ($<$.001) & 0.471 ($<$.001) \\
    \bottomrule
  \end{tabular}
  \caption{OLS regressions of overall impression change on each task's expectations-met rating, fitted separately for each task ($N{=}162$ per regression). Predictors and outcomes are z-scored.}
  \label{tab:appendix-pertask-exp}
\end{table}

\begin{table}[b]
  \centering
  \small
  \begin{tabular}{lcccc}
    \toprule
    \textbf{Task} & \textbf{GPT} & \textbf{Claude} & \boldmath{$t$} & \boldmath{$p$} \\
    \midrule
    Image generation & 2.97 & 2.92 & 0.49 & .628 \\
    Outreach message & 4.04 & 3.94 & 0.91 & .365 \\
    Acronym building & 3.93 & 3.72 & \textbf{3.08} & .002 \\
    \midrule
    Overall          & 3.76 & 3.61 & \textbf{2.81} & .006 \\
    \bottomrule
  \end{tabular}
  \caption{Mean judge-scored performance by model family, with independent-samples $t$-tests per task and for the overall composite.}
  \label{tab:appendix-family}
\end{table}

\clearpage

\end{document}